\documentclass{article}

\usepackage{microtype}
\usepackage{graphicx}
\usepackage{booktabs} 

\usepackage{tcolorbox}
\usepackage{enumitem}
\usepackage{url}
\usepackage{algorithmic}
\usepackage{multirow}%
\usepackage{amsfonts}%
\usepackage{mathrsfs}%
\usepackage[title]{appendix}%
\usepackage{textcomp}%
\usepackage{manyfoot}%
\usepackage{listings}%
\usepackage{anyfontsize}
\usepackage{color}
\usepackage{bbding}
\usepackage{longtable}
\usepackage{supertabular}
\usepackage{colortbl}
\usepackage{tocloft}
\usepackage{wrapfig} 
\usepackage{mathrsfs}
\usepackage{pifont} 
\usepackage{animate}

\definecolor{linkdarkblue}{rgb}{0,0.08,0.45}
\usepackage[
pdfpagelabels,bookmarksopen=true,hidelinks,
colorlinks=true,
allcolors=linkdarkblue,
]{hyperref}

\usepackage[accepted]{icml2025}

\usepackage{amsmath}
\usepackage{amssymb}
\usepackage{mathtools}
\usepackage{amsthm}

\usepackage[capitalize,noabbrev]{cleveref}

\theoremstyle{plain}

\theoremstyle{definition}

\theoremstyle{remark}

\usepackage[textsize=tiny]{todonotes}

\icmltitlerunning{Unveiling the Mystery of Weight in Large Foundation Models}

\begin{document}

\twocolumn[
\icmltitle{
         Unveiling the Mystery of Weight in Large Foundation Models: \\
         \textit{Gaussian Distribution Never Fades}
}


\icmlsetsymbol{equal}{*}

\begin{icmlauthorlist}
\icmlauthor{Chongjie Si}{sch}
\icmlauthor{Jingjing Jiang}{sch}
\icmlauthor{Wei Shen}{sch}

\end{icmlauthorlist}

\icmlaffiliation{sch}{MoE Key Lab of Artificial Intelligence, AI Institute, Shanghai Jiao Tong University.}

\icmlcorrespondingauthor{Chongjie Si}{chongjiesi@sjtu.edu.cn}
\icmlcorrespondingauthor{Wei Shen}{wei.shen@sjtu.edu.cn}

\icmlkeywords{Large Foundation Model, Weight Matrix, ICML}

\vskip 0.3in
]



\printAffiliationsAndNotice{}  

\begin{abstract}
This paper presents a pioneering exploration of the mechanisms underlying large foundation models' (LFMs) weights, aiming to simplify AI research.
Through extensive observation and analysis on prevailing LFMs, we find that regardless of initialization strategies, their weights predominantly follow a Gaussian distribution, with occasional sharp, inverted T-shaped, or linear patterns.
We further discover that the weights share the i.i.d. properties of Gaussian noise, and explore their direct relationship.
We find that transformation weights\footnote{Transformation weights denote the difference between pre-trained and fine-tuned weights during downstream task adaptation.} can be derived from Gaussian noise, and they primarily serve to increase the standard deviation of pre-trained weights, with their standard deviation growing with layer depth. 
In other words, transformation weights broaden the acceptable deviation from the optimal weights, facilitating adaptation to downstream tasks. 
Building upon the above conclusions, we thoroughly discussed the nature of optimal weights, ultimately concluding that they should exhibit zero-mean, symmetry, and sparsity, with the sparse values being a truncated Gaussian distribution and a few outliers.
Our experiments in LFM adaptation and editing demonstrate the effectiveness of these insights.
We hope these findings can provide a foundational understanding to pave the way for future advancements in the LFM community.
\end{abstract}

\section{Introduction}

Large foundation models (LFMs) \cite{qin2023chatgpt, touvron2023llama, devlin2018bert, kirillov2023segment} have exhibited remarkable performance across a diverse range of tasks \cite{sap2020commonsense, ma2024segment, rombach2022highstablediffusion}, and are widely regarded as a key pathway toward achieving Artificial General Intelligence \cite{feng2024far}.
However, their applications, including pretraining \cite{guu2020retrieval}, adaptation \cite{si2024see}, editing \cite{ilharco2022editing}, and compression \cite{zhu2023survey}, are fraught with intricate engineering challenges and resource limitations, making AI research challenging and thorny. 
This leads us to ponder: could we find a way to explore the properties or principles of optimal weights, radically simplify the landscape of AI research?

This paper represents the first exploration into the intrinsic mechanisms of LFMs' weights, aiming to offer profound insights into their behavior. 
Through extensive observation and analysis of the weights across prevailing LFMs in natural language processing (NLP), computer vision (CV), and multi-modal (MM), we arrive at a series of surprising and illuminating conclusions as follows:
\begin{itemize}
    \item Regardless of the initialization strategies, nearly all pre-trained weights conform to a Gaussian distribution, with a rare subset displaying sharp, inverted T-shaped, or even linear patterns.
    \item The weights exhibit i.i.d. properties similar to Gaussian noise, and transformation weights can be directly derived from Gaussian noise.
    \item Transformation weights primarily increase the standard deviation of the pre-trained weights, with their standard deviation growing with layer depth. 
    In essence, they increase the acceptable variance from the optimal weights to facilitate adaptation to downstream tasks. 
    \item The smaller differences in their standard deviations correlate with more similar model performance\footnote{This is a hypothesis that has been preliminarily validated.}, highlighting the potential to develop evaluation beyond only test data reliance.    
\end{itemize}

\begin{figure*}[!ht]
    \centering
    \includegraphics[width=\linewidth]{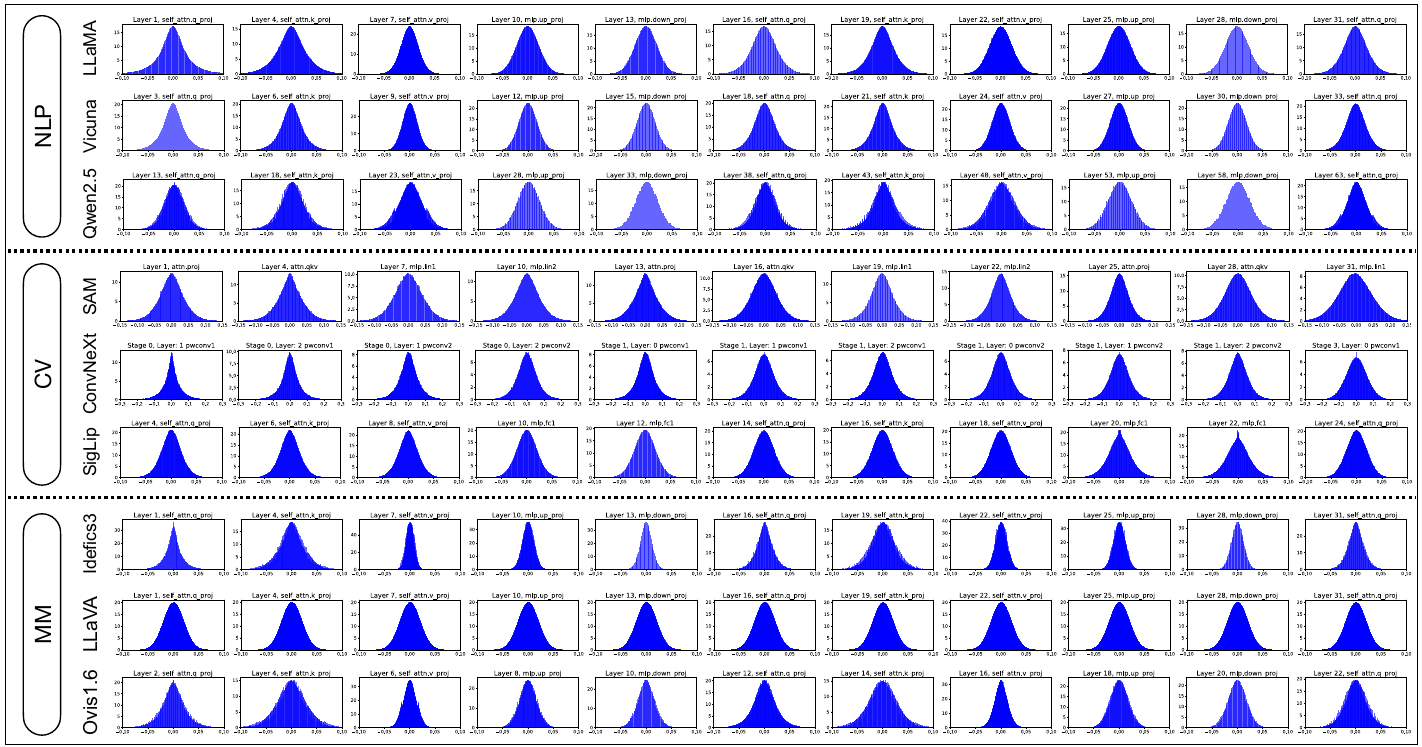}
    \caption{The distribution of pre-trained weights of prevailing large foundation models across NLP, CV, and MM. We show the weight distribution in different layers and modules. The distribution of these weights exhibits a remarkable resemblance to a Gaussian distribution. We randomly selected and showcased the distribution in several layers and modules. Additionally, we provide the weight distribution plots for each layer in the Appendix to offer a more comprehensive visualization.
}
    \label{fig:W distribution}
\end{figure*}
Building upon the aforementioned conclusions, we conducted an in-depth exploration of optimal weights and ultimately concluded that \textbf{optimal weights should exhibit zero-mean, symmetry, and sparsity, with sparse values likely following a truncated Gaussian distribution, interspersed with outliers}\footnote{Outliers refer to the elements that deviate significantly from the normal distribution.}.
We then extended these findings through further experiments, uncovering more profound relationships between these observations and the challenges faced by LFMs, leading to the following conclusions:
\begin{itemize}
    \item \textbf{Adaptation:} The essence of LFM adaptation lies in adjusting both Gaussian signals and outliers, specifically by adding the standard deviation of the pre-trained weights and updating the outliers to better align with task-specific requirements.
    \item \textbf{Editing:} The essence of LFM editing lies in selectively amplifying and preserving task-relevant outliers while smoothing the Gaussian signals to minimize interference, thereby enhancing the efficiency and robustness.
\end{itemize}

\begin{figure*}[!ht]
    \centering
    \includegraphics[width=\linewidth]{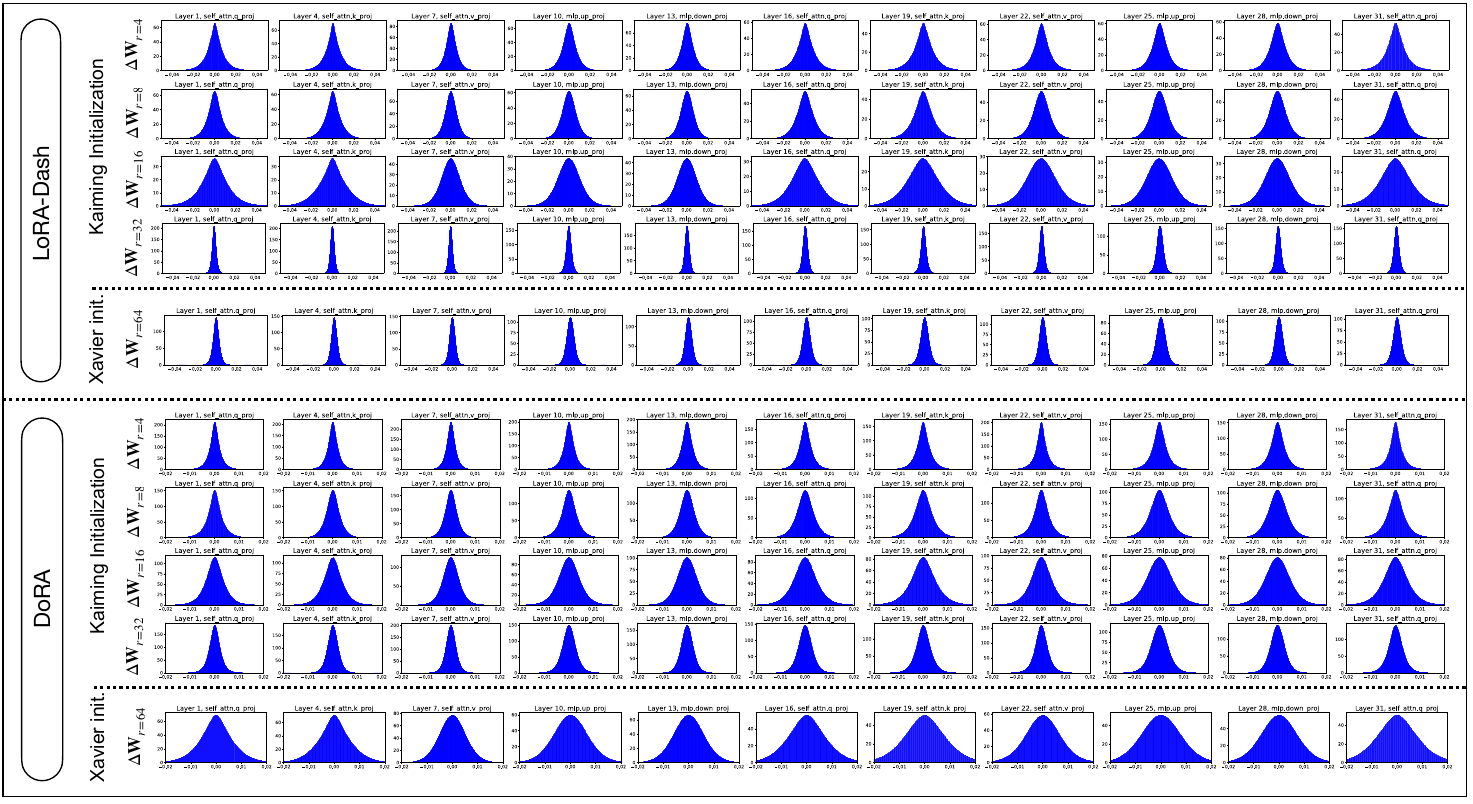}
    \caption{The distribution of the transformation matrices with different forms learned by two adaptation methods when fine-tuning LLaMA-7B on commonsense reasoning tasks, including different settings and initialization strategies. Clearly, regardless of different settings, initializations, or computation methods, the transformation weights closely resemble a Gaussian distribution. We randomly selected and showcased the distribution in several layers and modules. The weights distribution for each layer are shown in the Appendix.}
    \label{fig:Delta W distribution}
\end{figure*}

Furthermore, taking LFM adaptation and editing as concrete examples, we demonstrate the immense potential of observations.
Our method achieves performance gains of up to 8 points in adaptation and 2 points in editing compared to the baseline, demonstrating its effectiveness in these two domains.
We hope that this work could provide valuable insights and foster further advancements in the broader LFM community, finally simplify AI research\footnote{The contributions are listed in Appendix. \ref{sec conclusions}.}.

\section{Analysis on Weights Distribution}\label{sec analysis on weights distribution}

The prevailing consensus in most studies is that model weights encapsulate the knowledge learned from the data \cite{eilertsen2020classifying, schurholt2024towards}. 
Consequently, these weights exhibit distinct structural properties that reflect the underlying patterns captured by the model \cite{monasson1995weight, lee2023differentiable}.
Intuitively, it is widely accepted that the weights should be precise and resilient to perturbations, as they directly influence the model’s performance and generalization ability. 
However, what exactly do these weights look like? 

\subsection{Observation}

In the broader LFM community, the weights available for analysis can primarily be categorized into two types: pre-trained weights and fine-tuned weights obtained from adapting LFMs for downstream tasks.
To explore the properties of weights, we first examine the distributions of these two categories.
Since the latter can be viewed as a specific case of the former, we diversify our analysis by focusing on the transformation weights $\Delta\mathbf{W}$, which represents the changes required to adapt the pre-trained weights $\mathbf{W}$ to a downstream task\footnote{The fine-tuned weights can be expressed as $\mathbf{W} + \Delta \mathbf{W}$. Besides, $\Delta\mathbf{W}$ can be directly obtained through various fine-tuned methods, such as LoRA \cite{hu2021lora}.}. 
For the pre-trained weights, we choose nine of the latest and most prominent LFMs from NLP, CV, and MM, including LLaMA-7B \cite{touvron2023llama}, Vicuna-13B \cite{zheng2023judgingvicuna}, Qwen2.5-32B \cite{qwen2.5}, SAM-h \cite{kirillov2023segment}, ConvNeXt-xlarge \cite{woo2023convnext}, SigLip \cite{zhai2023sigmoid}, Idefics3-8B \cite{laurenccon2024buildingidefics3}, LLaVA-NeXT \cite{liu2024llavanext}, and Ovis1.6 \cite{lu2024ovis}\footnote{Details can be found in Appendix. \ref{sec supp detail lfm}.}. 
As for the transformation weights, we utilized weights obtained by fine-tuning LLaMA-7B \cite{touvron2023llama} on commonsense reasoning tasks using LoRA-Dash \cite{si2024unleashing} and DoRA \cite{liu2024dora}\footnote{Details can be found in Appendix. \ref{sec supp detail lfm}.}.

\begin{table*}[!ht]
 \renewcommand\arraystretch{0.8}
 \setlength{\tabcolsep}{1.1mm}
    \centering
    \caption{Average skewness and Kurtosis of the weights in different LFMs. }
    \resizebox{\textwidth}{!}{
    \begin{tabular}{ c c c c c c c c c c  }
    \toprule
    Metric & LLaMA-7B & Vicuna-13B & Qwen2.5-32B & SAM-h & ConvNeXT-xlarge & SigLip & Idefics3-8B & LLaVA-NeXT & Ovis1.6\\ \midrule
    Ratio &  99.51\% & 99.49\% & 99.36\% & 99.26\% & 99.18\% & 99.39\% & 99.39\% & 99.73\% & 99.40\% \\
    
    Skewness & 0.00$_{\pm0.00}$ & 0.00$_{\pm0.00}$ &  0.00$_{\pm0.00}$ & 0.00$_{\pm0.00}$ & 0.00$_{\pm0.00}$ & 0.00$_{\pm0.00}$ & 0.00$_{\pm0.00}$ & 0.00$_{\pm0.00}$ & 0.00$_{\pm0.00}$ \\
    
    Kurtosis & 3.02$_{\pm0.24}$ & 3.06$_{\pm0.44}$ & 3.27$_{\pm1.32}$ & 3.26$_{\pm0.47}$ & 3.33$_{\pm0.44}$ & 3.12$_{\pm0.38}$ & 3.12$_{\pm0.36}$ & 2.83$_{\pm0.00}$ & 3.10$_{\pm0.22}$\\
    \bottomrule
    \end{tabular}}
    \label{tab:skewness and kurtosis}
\end{table*}

\begin{figure*}[!ht]
    \centering
    \includegraphics[width=0.92\linewidth]{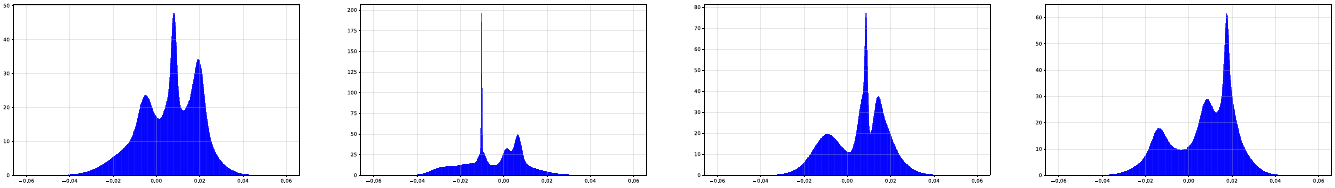}
    \caption{The distribution of the elements if they are independent but not identically distributed. The subfigures represent the overall distributions derived under the assumption that each element follows a different Gaussian distribution.}
    \label{fig:random gaussian}
\end{figure*}

The distributions of these weights are illustrated in Figs. \ref{fig:W distribution}-\ref{fig:Delta W distribution}.
Surprisingly, we observed that across different models, layers, and modules, the pre-trained weights exhibit a striking resemblance to Gaussian signals with a mean of almost zero (smaller than $10^{-5}$).
Additionally, for the transformation weights derived from LoRA-Dash and DoRA, we reached the same conclusion regarding their Gaussian-like distribution. 
Furthermore, considering that most existing model training relies on Kaiming initialization \cite{he2015delving}, we also employed different initialization strategies (i.e., Xavier initialization \cite{glorot2010understandingtheory3}) to eliminate the potential influence of initialization in LoRA-Dash and DoRA. 
However, the results are strikingly consistent: the derived weights still exhibited a Gaussian distribution.

Given that weights are believed to encapsulate knowledge, this finding suggests that Gaussian signals might inherently encode the knowledge learned from data. 
Even more intriguingly, we found that the weights of multi-modal models also follow a Gaussian distribution, implying that such distributions are capable of carrying information across multiple modalities...But wait—do the weights truly follow a Gaussian distribution?

\subsection{Validation}
To determine whether the weight distributions indeed resemble a Gaussian distribution, we adopt a multi-step statistical analysis.
First, for the elements in each matrix, we calculated their mean ($\mu$) and standard deviation ($\sigma$). 
To mitigate the impact of outliers, we filter the elements by retaining the values within the range $[\mu - 3\sigma, \mu + 3\sigma]$\footnote{In a Gaussian distribution, approximately 99.7\% of the data lies within the 3$\sigma$ range, while 95.4\% of the data lies within the 2$\sigma$ range, according to the empirical \textit{68–95–99.7 rule}.}. 
Indeed, we observed that almost all $\mu$ of the weight elements is approximately zero. However, to account for potential errors or deviations, we continue to use $\mu$ in the ranges.
Alongside this filtering process, we computed a ratio indicating the proportion of retained points, which reflects the quality of the weight distribution and its concentration around it.

Next, we evaluated the distributions using two key metrics: skewness and kurtosis\footnote{Please refer to Appendix \ref{sec supp detail skew and kurtosis} for more details.} \cite{joanes1998comparing, decarlo1997meaning, bulmer2012principles}. 
Specifically, skewness measures the asymmetry of a distribution, and kurtosis describes the ``tailedness'' of a it, quantifying how heavy or light the tails are compared to a Gaussian distribution. 
For a perfectly symmetric Gaussian distribution, the skewness is theoretically 0, and the kurtosis (non-Fisher adjusted) is 3.

We computed the mean and standard deviation of these two metrics—skewness and kurtosis—across all layers and modules of the aforementioned models.
Additionally, we report the ratio of retained data points used for distribution analysis. 
The results are presented in Tables \ref{tab:skewness and kurtosis} and \ref{tab:skewness and kurtosis transformation} in Appendix. 
It is evident that, within a small margin of error, the weight distributions of these models align closely with Gaussian properties. 
The skewness values are consistently 0 across all models, indicating symmetric distributions, while the kurtosis values are near the Gaussian reference of 3, with minor variations among certain models. 
These observations strongly support that the weights follow a Gaussian distribution. 
Furthermore, a broader review of studies in different fields reveals that similar Gaussian-like phenomena have been noted in weight distributions \cite{huang2024billm, jie2023revisiting}.
This consistency leads us to conclude that \textbf{the weights in LFMs truly follow a Gaussian distribution}.

\section{Derivation from Gaussian Noise}

\subsection{I.i.d Properties of Weights}

Moreover, some studies \cite{thamm2022random,anonymous2024loca, yang2020tensor,lee2017deep} suggest that the individual elements of a weight matrix, denoted as $\{w_i\}_{i=1}^n$, are asymptotically independent. 
Assume that each element $w_i$ follows a certain distribution $\mathcal{P}_i(w)$.
The overall distribution of the weights is then determined by the specific forms of these individual distributions.
If the distribution of each weight element $w_i$ is distinct, that is, $\mathcal{P}_i(w) \neq \mathcal{P}_j(w)$ for some $i \neq j$, then the combined distribution of the weights would not form a single Gaussian distribution, as illustrated in Fig. \ref{fig:random gaussian}.
However, based on our exploration in the previous section, we observed that the overall weight distributions in LFMs are Gaussian distributions. 
Given that $w_i$ are independent, their joint distribution can be expressed as the product of their marginal distributions, i.e.,
\begin{equation}
\begin{aligned}
    \mathcal{P}(\{w_i\}_{i=1}^n) &= \frac{1}{(\sqrt{2\pi\sigma^2})^{n}}\exp\left(-\frac{1}{2\sigma^2}\sum_{i=1}^n (w_i-\mu)^2\right) \\
    & = \prod_{i=1}^n \frac{1}{\sqrt{2\pi\sigma^2}}\exp\left(-\frac{1}{2\sigma^2} (w_i-\mu)^2\right)\\
    &= \prod_{i=1}^n \mathcal{P}_i(w).
\end{aligned}
\end{equation}
This indicates that the independent weight elements must share the same underlying distribution, i.e.,
\begin{equation}
    w_i \sim \mathcal{P}_i(w)=\mathcal{N}(\mu, \sigma^2), \quad \forall i.
\end{equation}
Therefore, the weight elements are asymptotically independent and identically distributed (i.i.d.).

Each element being independently and identically distributed, combined with the fact that the overall distribution adhering to a Gaussian distribution, inevitably brings to mind the characteristics of Gaussian noise. 
This naturally leads to an unexpected yet intuitive question: Can weights be directly derived from Gaussian noise?

\begin{table*}[ht]
    \centering
    \renewcommand\arraystretch{0.6} 
    \caption {Results with DeBERTaV3-base \cite{he2021debertav3} fine-tuned on GLUE development set.}
    \resizebox{\textwidth}{!}{
    \begin{tabular}{c | c | c| c c c c c c c c |
    c}
    \toprule
        \multirow{2}{*}{\textbf{Row}} & \multirow{2}{*}{\textbf{Method}} & \multirow{2}{*}{\# \textbf{Params}} & \textbf{MNLI} & \textbf{SST-2} & \textbf{CoLA} & \textbf{QQP} & \textbf{QNLI} & \textbf{RTE} & \textbf{MRPC} & \textbf{STS-B} & \textbf{All} \\
        &  & & Acc & Acc & Mcc & Acc & Acc & Acc & Acc & Corr & Avg. \\ \midrule

         1 & Head & 0 & 65.37 & 83.72 & 52.93 & 78.37 & 72.43 & 62.82 & 75.74 & 82.12 & 71.69\\
         
         2 & Fully FT & 184M & 89.90 & 95.63 & 69.19 & 92.40 & 94.03 & 83.75 & 89.46 & 91.60 & 88.24 \\ \midrule


        3 & SVDiff & 55296 & 68.32 & 84.28 & 55.21 & 81.91 & 75.86 & 66.06 & 76.47 & 82.82 & 73.87 \\
        
        4 & Ours & 72 & 69.49 & 88.53 & 61.34 & 84.02 & 83.58 & 71.12 & 83.58 & 87.29 & 78.62 \\

        \midrule


        
        5 & Fully FT & Gaussian & 35.45 & 66.97 & 7.11 & 64.86 & 57.04 & 52.35 & 69.36 & 8.91 & 45.13 \\
        
        
        \bottomrule
        
    \end{tabular}
    }
    \label{tab:deberta results}
\end{table*}

\subsection{Validation}\label{sec weight noise validation}

In a statistical sense, noise is a relative concept, referring to information that is not helpful for the current task. 
Notably, what may be perceived as noise for one task could be crucial for another, effectively becoming ``weights'' in that context.
This duality leads us to believe that weights and noise are fundamentally indistinguishable in their structure and behavior.
To validate this hypothesis, we begin with transformation weights, as they are computationally inexpensive to train and readily accessible.

We aim to achieve adaptation of the pre-trained model DeBERTaV3-base \cite{he2021debertav3} on the GLUE benchmark \cite{wang2018glue}, which consists of eight datasets for natural language understanding (NLU) tasks. 
For a pre-trained weight matrix $\mathbf{W}$, we aim to learn the transformation matrix $\Delta \mathbf{W}$ that represents the updates required for adaptation \cite{hu2021lora,si2024flora}. 
To significantly reduce the number of trainable parameters, we adopt a ``cheating'' approach by leveraging auxiliary information (e.g., information of fully fine-tuned weights) to pre-construct $\Delta\mathbf{W}$. 
During the training phase, we only train a scalar coefficient $s$ while keeping $\mathbf{W}$ and $\Delta\mathbf{W}$ frozen, such that the fine-tuned weight matrix $\mathbf{W}'$ is represented as:
\begin{equation}
    \mathbf{W}' = \mathbf{W} + s \Delta\mathbf{W}.
    \label{eq:gaussian knowledge}
\end{equation}
The results are presented in Table \ref{tab:deberta results}. 
For comparison, we include the results of training only the head (row 1) and fully fine-tuning the model (row 2). 
We also include the results of a efficient adaptation method SVDiff \cite{han2023svdiff}.
First, comparing the results in rows 1 and 4 demonstrates that by training only 72 parameters, we achieve performance that significantly surpasses the Head baseline. 
When comparing rows 3 and 4, it is obvious that we train only a tiny fraction of the parameters in SVDiff (over 700 times fewer), yet achieve superior performance.
This indicates that the learned parameters significantly capture more task-specific knowledge.
Moreover, when comparing rows 2 and 4, on certain datasets, the results obtained by training these 72 parameters (a mere 0.00004\% of the total parameters) even approach the performance of fully fine-tuning.
Considering the extremely small number of trainable parameters and the remarkable performance achieved, it is reasonable to attribute this success to the vast amount of task-specific knowledge embedded in the pre-constructed $\Delta \mathbf{W}$, which indeed should not be available during standard training. 

\textit{However, what if we told you that these carefully crafted $\Delta \mathbf{W}$ matrices were, in fact, randomly matrices using standard Gaussian noise?}

For each pre-trained weight matrix $\mathbf{W}$, we did not design the structure of $\Delta \mathbf{W}$ or rely on any external information at all. 
Instead, we randomly initialized $\Delta \mathbf{W}$ with standard Gaussian noise and, during the training phase, adjusted only the standard deviation of the Gaussian (i.e., the scalar parameter $s$ in Eq. (\ref{eq:gaussian knowledge})). 
Additionally, it is important to note that the results presented are the average of five runs. 
It means that the Gaussian noise generated in each run is different, further suggesting that transformation weights exhibit randomness and independence.
When not informed, one might perceive $\Delta \mathbf{W}$ as meaningful weights, or even consider it carefully designed; yet upon closer inspection, it is revealed that $\Delta \mathbf{W}$ can simply derived from Gaussian noise with a specific standard deviation.
This shows a similarity between transformation weights and Gaussian noise in both structure and behavior, leading us to the conclusion: \textbf{transformation weights can be directly derived from Gaussian noise.}

However, why transformation weights can be derived from Gaussian noise?
Why adding such ``noise'' can improve the model performance?
When we revisit this part after exploring the next, we may find ourselves marveling at the layers of complexity that could redefine our understanding of their intricate interplay.

\subsection{Revisiting the Transformation Weight Distribution}

As Isaac Newton once remarked, ``Nature is pleased with simplicity, and affects not the pomp of superfluous causes'' \cite{newton1930principia}, we believe that the oracle weights should adhere to this principle of simplicity. 
Consider an optimal weight matrix $\mathbf{W}^*$ derived from an immensely large dataset $\mathcal{D}$. 
To simplify notation, we treat the matrix as being flattened.
During the pre-training phase\footnote{Note that fine-tuning can also be viewed as a kind of pre-training, and vice versa.}, we sample $n$ examples from $\mathcal{D}$ and train the model by optimizing a loss function to obtain the weight matrix $\mathbf{W}$.
Consequently, $\mathbf{W}$ can be viewed as an M-estimator of $\mathbf{W}^*$ \cite{van2000asymptotic,peracchi1990robust}. 
Under fairly general conditions, the difference $\mathbf{W} - \mathbf{W}^*$ is known to be asymptotically normal \cite{yohai1979asymptotic, hoeffding1992class}, i.e., 
\begin{equation}
    \sqrt{n} (\mathbf{W} - \mathbf{W}^*) \xrightarrow{d.} \mathcal{N}(0, \sigma^2 \mathbf{I}).
\end{equation}
In other words, 
\begin{equation}
    \mathbf{W} = \mathbf{W}^* + \mathcal{N}(0, \frac{\sigma^2}{n}\mathbf{I}) + o(\frac{1}{\sqrt{n}}).
    \label{eq w w* gaussian}
\end{equation}
In practice, the third term can be assumed to be negligible when the number of training samples is sufficiently large. 
The second term represents the noises arisen during training \cite{zhou2019toward,smith2020generalization}.
This indicates that the obtained weight matrix $\mathbf{W}$ can be interpreted as a sample drawn from a Gaussian distribution, where the mean is the optimal weight matrix $\mathbf{W}^*$, and the variance represents the range of acceptable variation from the optimal weight.

We here explain why transformation weights can be derived from Gaussian noise and why merely adding such noise can improve overall performance in Sec. \ref{sec weight noise validation}. 
We could consider both the pre-trained and fine-tuned weights as samples drawn from $\mathbf{W}^*$, as the datasets for training these weights are the subsets of that for training $\mathbf{W}^*$. 
Since the transformation weight represents the difference between these two, it naturally forms a Gaussian distribution with a mean of zero\footnote{Please refer to Appendix. \ref{sec supp weight delta remark} for more details.}.
Moreover, since the number of training samples for fine-tuned weights is generally much smaller than that for pre-training, the corresponding Gaussian noise for sampling $\mathbf{W}^*$ has a larger standard deviation, as suggested by Eq. (\ref{eq w w* gaussian}).
Therefore, adding Gaussian noise to the pre-trained weights essentially increases this standard derivation when sampling $\mathbf{W}^*$, corresponding to learning the fine-tuned weights.

We thus can interpret the scalar $s$ learned in Eq. (\ref{eq:gaussian knowledge}) as compensating for the standard deviation of the Gaussian noise. 
To further validate this, we analyzed the difference in $s$\footnote{Calculate as $| |s_1| -|s_2||$ for two different scalars.} obtained from two different seeds in the experiments of Sec. \ref{sec weight noise validation}.
We found that the difference in $s$ is consistently small across different datasets, with the average error 0.009\footnote{For each dataset, we averaged the differences of four groups (5 seeds) across all layers, then averaged the results across datasets.}.
This finding further demonstrates that even when $\Delta\mathbf{W}$ is initialized with different standard Gaussian noise, the trained $s$ values differ minimally, confirming that $s$ indeed serves to adjust for the standard deviation of the Gaussian noise. 
This also explains why different Gaussian noise pre-constructed $\Delta\mathbf{W}$ can consistently achieve superior performance.

Moreover, considering that the Gaussian noise corresponding to $\Delta \mathbf{W}$ must satisfy a specific standard deviation, this implicitly imposes certain structural requirements.
We posit that this structure is what allows the weights to function as ``weights''. 
However, it seems that this structure may be far from sufficient—while increasing the Gaussian standard deviation improves performance by nearly 7 points over the baseline, it still lags 10 points behind fully fine-tuning, which is a seemingly negligible gap. 
Why does this disparity exist?
With this question in mind, we turn our attention to the pre-trained weights, only to find that things become far more nuanced.


\section{Unveiling the Mystery of Weight}

Our observations reveal that the pre-trained weights follow a zero-mean Gaussian distribution. 
Based on Eq. (\ref{eq w w* gaussian}), it is straightforward to deduce that the optimal weights are also a zero-mean Gaussian.
Philosophically, we posit that $\mathbf{W}^*$ embodies simplicity and highly compressibility.
A Gaussian distribution indeed appears to satisfy these assumptions.

To validate this, we initialized DeBERTaV3-base with a standard Gaussian distribution and subsequently trained only its corresponding standard deviation. 
As a result, the fine-tuned weights obtained at the end strictly follow a Gaussian distribution.
However, as shown in row 5 in Table \ref{tab:deberta results}, the model’s performance is only about half of that achieved with full training.
This comes to the conflict: if the optimal weights were indeed a Gaussian distribution, then the model results should, as expected, be superior.

When theory and experiments conflict, we need to revisit our entire line of reasoning. 
Eventually, we identified a critical assumption: the Gaussian distribution of $\mathbf{W}$ was derived by focusing exclusively on statistical values within the 3$\sigma$ range, discarding outliers with exceptionally large magnitudes.
We then conducted extensive literature research on outliers and discovered that some studies highlight how outliers can influence model outputs \cite{yin2023outlier, kovaleva2021bert,puccetti2022outliers}. 
\cite{yadav2024ties} even demonstrates that retaining only the outliers in the weights can achieve more than 70\% of the model’s original performance.
This motivates us to rethink the distribution of weights, which leads to the following conclusions:

\begin{table*}[!ht]
 \renewcommand\arraystretch{0.8}
 \setlength{\tabcolsep}{1.8mm}
    \centering
    \caption{Results on commonsense reasoning tasks. We fine-tune LLaMA-7B, LLaMA2-7B and LLaMA3-8B on this task.}
    \resizebox{\textwidth}{!}{
    \begin{tabular}{c | c c | c c c  c c c c c | l }
    \toprule
    \textbf{Model} & \textbf{Method} & \textbf{Params(\%)} & \textbf{BoolQ} & \textbf{PIQA} & \textbf{SIQA} & \textbf{HellaS.} & \textbf{WinoG.} & \textbf{ARC-e} & \textbf{ARC-c} & \textbf{OBQA} & \multicolumn{1}{c}{\textbf{Avg.}} \\
    \toprule
    ChatGPT & - & - & 73.1 & 85.4 & 68.5 & 78.5 & 66.1 & 89.8 & 79.9 & 74.8 & 77.0 \\ \midrule
    
    \multirow{4}{*}{LLaMA-7B} & LoRA$_{r=16}$ & 0.42 & 69.9 & 77.8 & 75.1 & 72.1 & 55.8 & 77.1 & 62.2 & 78.0 & 70.9 \\ 
    
     & \cellcolor{gray!20}LoRA+Ours & \cellcolor{gray!20}0.42 & \cellcolor{gray!20}68.1 & \cellcolor{gray!20}81.8 & \cellcolor{gray!20}77.2 & \cellcolor{gray!20}84.2 & \cellcolor{gray!20}71.3 & \cellcolor{gray!20}78.2 & \cellcolor{gray!20}62.3 & \cellcolor{gray!20}77.4 & \cellcolor{gray!20}75.1 (\textcolor[rgb]{0.25,0.5,0.75}{+4.2}) \\ \cline{2-12}
     
    & LoRA$_{r=32}$ & 0.83 & 68.9 & 80.7 & 77.4 & 78.1 & 78.8 & 77.8 & 61.3 & 74.8 & 74.7 \\  \rowcolor{gray!20}

    \cellcolor{white} & LoRA+Ours & 0.83 & 69.5	& 82.3 & 78.1 & 80.5 & 81.8 & 81.5 & 65.6 & 79.0 & 77.3 (\textcolor[rgb]{0.25,0.5,0.75}{+3.6})  \\

    \midrule

    \multirow{4}{*}{LLaMA2-7B} & LoRA$_{r=16}$ & 0.41 &  71.2 & 83.0 & 68.0 & 72.1 & 80.9 & 73.3 & 59.0 & 71.8 & 72.4 \\ 

    & \cellcolor{gray!20}LoRA+Ours & \cellcolor{gray!20}0.41 & \cellcolor{gray!20}72.0 &  \cellcolor{gray!20}82.5 & \cellcolor{gray!20}79.3 & \cellcolor{gray!20}89.1 & \cellcolor{gray!20}83.0 & \cellcolor{gray!20}83.3 & \cellcolor{gray!20}70.2 & \cellcolor{gray!20}82.0 & \cellcolor{gray!20}80.2 (\textcolor[rgb]{0.25,0.5,0.75}{+7.8}) \\ \cline{2-12}

    & LoRA$_{r=32}$ & 0.82 & 69.1 & 80.4 & 78.0 & 80.6 & 81.1 & 80.0 & 66.5 & 79.2 & 76.8 \\ 
    
    \rowcolor{gray!20}

    \cellcolor{white} & LoRA+Ours & 0.82 & 71.6 & 83.1 & 79.5 & 85.8 & 82.2 & 81.2 & 66.4 & 78.8 & 78.6 (\textcolor[rgb]{0.25,0.5,0.75}{+1.8})  \\
     
    \midrule

    \multirow{4}{*}{LLaMA3-8B} & LoRA$_{r=16}$ & 0.35 & 72.3 & 86.7 & 79.3 & 93.5 & 84.8 & 87.7 & 75.7 & 82.8 & 82.8 \\

    & \cellcolor{gray!20}LoRA+Ours & \cellcolor{gray!20}0.35 & \cellcolor{gray!20}72.6 & \cellcolor{gray!20}87.9 & \cellcolor{gray!20}79.9 & \cellcolor{gray!20}94.4 & \cellcolor{gray!20}85.8 & \cellcolor{gray!20
    }87.8 & \cellcolor{gray!20}76.6 & \cellcolor{gray!20}84.6 & \cellcolor{gray!20}83.7 (\textcolor[rgb]{0.25,0.5,0.75}{+0.9})  \\
    
    \cline{2-12}

    & LoRA$_{r=32}$ & 0.70 & 70.8 & 85.2 & 79.9 & 91.7 & 84.3 & 84.2 & 71.2 & 79.0 & 80.8 \\ \rowcolor{gray!20}

    \cellcolor{white} & LoRA+Ours & 0.70 & 71.7 & 86.2 & 80.1 & 93.3 & 84.5 & 87.0 & 74.4 & 84.4 & 82.7 (\textcolor[rgb]{0.25,0.5,0.75}{+1.9}) \\
    
    \bottomrule
    \end{tabular}}
    \label{tab:results of commonsense}
\end{table*}

\textit{We hypothesize that the optimal weight $\mathbf{W}^*$ is a sparse matrix containing a few values (e.g., outliers, Gaussian signals, etc).
Under current pre-training techniques, we obtain its M-estimator, the pre-trained weights $\mathbf{W}$. 
Due to dataset limitations, $\mathbf{W}$ includes some values in $\mathbf{W}^*$ that are closely related to the pre-training dataset while neglecting others. 
Combined with Gaussian noise introduced during training, this forms the observed distribution of pre-trained weights\footnote{Please refer to Appendix. \ref{supp derivation weight} for more details.}.
Moreover, downstream tasks may involve values in $\mathbf{W}^*$ that are critical to those tasks but absent from the pre-trained weights.
Consequently, the transformation matrix cannot solely consist of Gaussian signals; it must also incorporate a combination of Gaussian signals and some values in $\mathbf{W}^*$. 
This also aligns closely with our observations.}

Unfortunately, due to page limitations, we are unable to fully present all of our findings in the main-text, but \textcolor{red}{\textbf{we strongly recommend readers to refer to Appendixes. \ref{supp weight distribution}-\ref{sec supp open discussion}}} for more detailed explorations and discussions on the weight distribution and mysterious $\mathbf{W}^*$, including 1. The closer the standard deviations of two $\Delta \mathbf{W}$, the more similar the corresponding model performance becomes. 
2. The standard deviation of $\Delta \mathbf{W}$ increases with the layer depth.
Furthermore, we provide a detailed analysis of $\mathbf{W}^*$, deriving the reasons behind the observed weight distribution and offering a more fundamental explanation for multiple methods.

With this, our exploration in this work comes to an end.
However, our investigation is far from over, and further research will be presented in the future.

\section{Application}

One might wonder about the potential value and practical application of our exploration of weight matrices.
First, our observations provide a solid explanation for the rationale behind many existing approaches in various fields, which we believe is a significant contribution. 
Moreover, based on our detailed exploration of $\Delta \mathbf{W}$, without preamble, we turn to two representative applications: Parameter-Efficient Fine-Tuning (\textit{adaptation}), and Model Merging (\textit{editing}).

\subsection{Parameter Efficient Fine-tuning}
Parameter-efficient fine-tuning (PEFT) aims to learn fewer parameters while achieving comparable performance compared to fully fine-tuning.
In the previous sections, we established the intrinsic relationship between Gaussian noise and the transformation matrix during training.
Building upon this insight, we propose a plug-and-play augmentation method applicable to any existing approach: augment the pre-trained weight matrix $\mathbf{W}$ with a randomly initialized Gaussian noise term, while learning the standard deviation of this noise during training, as described in Eq. (\ref{eq:gaussian knowledge}). 
It not only effectively increases the standard deviation of Gaussian noise in the pre-trained weights but also simplifies the learning process for other methods, allowing them to focus on identifying and refining better outliers.

Furthermore, since the pre-trained weights $\mathbf{W}$ themselves follow a Gaussian distribution, they can be used as an initialization for Gaussian noise. 
We can directly initialize the transformation matrix using the pre-trained $\mathbf{W}$.
By doing so, the task reduces to learning the scaling factor $s$ for $\mathbf{W}$, thereby streamlining and optimizing the training process. 
Taking a prevailing PEFT method LoRA \cite{hu2021lora} as an instance, the weight update can be redefined as:
\begin{equation}
    \mathbf{W}' = (s + 1)\mathbf{W} + \mathbf{A}\mathbf{B},
\end{equation}
where $\mathbf{A}$ and $\mathbf{B}$ are two low-rank matrices introduced by LoRA.
For more details on PEFT and related methods like LoRA, please refer to Appendix \ref{sec related peft}.

The experimental results, as presented in the Table. \ref{tab:results of commonsense}, clearly demonstrate that LoRA combined with specific Gaussian noise significantly outperforms standalone LoRA.
Notably, this improvement is observed under both parameter configurations and for all three LLaMA variants.
By leveraging the duality of Gaussian noise, we provide a novel perspective for designing methods in PEFT, which further validates the reliability of our observations. 
For experimental details, please refer to Appendix \ref{sec supp detail cr task}

\begin{table*}[ht!]
 \renewcommand\arraystretch{0.4}
 \setlength{\tabcolsep}{1.3mm}
    \centering
    \caption{Results on model merging tasks. We merge two LLaVA-v1.5-13B \cite{liu2023llava} fine-tuned model: Instruction-tuned LLaVA-1.6-13B \cite{liu2023llava}, and Math fine-tuned Math-LLaVA \cite{shi2024math}. The results of LLaVA-v1.5-13B are cited from \cite{duan2024vlmevalkit}.}
    \resizebox{\textwidth}{!}{
    \begin{tabular}{c c | c c c c c c c | l}
        \toprule
        \textbf{Model} & \textbf{Type} & \textbf{MathVista} & \textbf{MMStar} & \textbf{MMMU} & \textbf{WeMath} & \textbf{AI2D} & \textbf{DynaMath} & \textbf{GeoQA} & \textbf{Avg} \\ 
        \midrule
        LLaVA-v1.5-13B & Pre-trained & 34.3 & 37.0 & 27.7 & - & 61.1 & - & - & - \\
        
        \midrule
        LLaVA-1.6-13B & 8bit, Instruction & 33.6 & 40.2 & 42.6 & 30.1 & 67.9 & 20.3 & 23.9 & 36.9 \\ 
       
        Math-LLaVA & 8bit, Math & 45.8 & 42.8 & 42.0 & 33.9 & 66.7 & 22.9 & 46.6 & 43.0 \\ 

         \midrule
         
        Average & &  43.7 & 42.5 & 43.0 & 35.2 & 69.3 & 22.6 & 41.2 & 42.5\\ 
        
      \rowcolor{gray!20}
      
         Ours  & $t=3$ & 44.4 & 42.3 & 44.0 & 38.2 & 71.2 & 24.7 & 41.4 & 43.7 (\textcolor[rgb]{0.25,0.5,0.75}{+1.2}) \\ 
         
         \rowcolor{gray!20}

         Ours & $t=2$ & 46.0 & 43.5 & 44.0 & 38.4 & 70.1 & 25.6 & 44.2 & 44.5 (\textcolor[rgb]{0.25,0.5,0.75}{+2.0}) \\ 
         

        \bottomrule
    \end{tabular}
    }
    \label{tab model merging}
\end{table*}

\subsection{Model Merging}
Model merging techniques aim to integrate multiple task-specific models into a single, cohesive model, retaining the strengths of each individual model while eliminating the need for access to the original training data.
Despite the various methods proposed for model merging, the dominant approach in large model technical reports remains averaging the weights of different models \cite{he2024efficient,team2024gemma,baichuan2023baichuan2}, due to its simplicity and effectiveness. 
Please refer to Appendix \ref{sec related model merging} for more details.

Based on our observations, we believe that the averaging process essentially performs a form of ``smoothing'' on the weights.
However, this averaging can also cause outliers to diminish, potentially reducing them to the same level as Gaussian signals. 
Therefore, we propose an intuitive approach: to amplify the outliers while averaging the others.

Firstly, to merge $n$ task-specific models, following \cite{yadav2024ties,ilharco2022editing}, we begin by obtaining the change in weights $\{\Delta\mathbf{W}^k_i|_{k=1}^p\}_{i=1}^n$ for each trained task-specific model relative to the pre-trained weights (i.e., the task vectors \cite{ilharco2022editing}).
Here, $p$ denotes the number of distinct parameter groups (i.e., a matrix or vector) in the model, and $\Delta \mathbf{W}^k_i$ represents the $k$-th group of weight changes for the $i$-th task-specific model.
Next, considering that a value which appears as noise in one model might be an outlier in another, for each parameter group $k$, we first compute the standard deviation $\sigma^k_i$ of the $k$-th parameter group for each of the $n$ models. 
Then, we determine the minimum standard deviation across all models for the $k$-th parameter group, defined as $\sigma^k = \min\{\sigma^k_1, \sigma^k_2, \cdots, \sigma^k_n\}$.
Subsequently, for each $\Delta \mathbf{W}^k_i|_{i=1}^n$, we classify its values into two categories: outliers and noises, based on a threshold $t\sigma^k$ ($t \in \{2, 3\}$)\footnote{Note that the mean of most weights is typically less than $10^{-5}$.}.
Specifically, values outside the range $[- t\sigma^k, t\sigma^k]$ are treated as outliers and remain unchanged. Values within this range are treated as noises and are averaged across the $n$ task-specific models. 
Finally, the processed weight changes $\{\Delta \mathbf{W}^k_i\}_{i=1}^n$ are summed and added back to the pre-trained weights to obtain the final merged model. 
The pseudo codes are shown in Alg. \ref{alg:model merging}.

The results in Tab. \ref{tab model merging} demonstrate the effectiveness of our approach.
Compared to the averaging method, selectively amplifying outliers beyond the 3$\sigma$ range significantly improves the merged model’s performance across all datasets.
Moreover, progressively amplifying more outliers (e.g., $t=2$) yields further performance gains, surpassing the traditional method by 2 points on average.
These findings not only highlight the superiority of our approach, but they also underscore the critical role that outliers play in shaping the performance of the merged model.
For experimental details, please refer to Appendix \ref{sec supp detail mm task}.

\section{Conclusion and Future Work}

In this work, we explored the underlying mechanisms of LFM weights, offering a deeper understanding of their structure and dynamics. 
Through comprehensive analysis and observation of existing LFMs, we identify that the weights primarily conform to a Gaussian distribution, though some exhibit atypical patterns such as sharp, inverted T-shaped, or linear forms.
We further reveal that these weights exhibit the i.i.d. properties of Gaussian noise, and explore the direct correlation between them.
Transformation weights, which can be derived from Gaussian noise, play a key role in increasing the standard deviation of the pre-trained weights, with their standard deviation increasing as layer depth grows.
Building on these findings, we conduct an in-depth examination of optimal weights and conclude that they should possess zero mean, symmetry, and sparsity, with sparse values likely following a truncated Gaussian distribution and featuring a few outliers.
Furthermore, we hypothesize and preliminarily validate that the closer the standard deviations of transformation weights, the more analogous the model performance becomes.
Our experiments in LFM adaptation and editing validate the potential of these findings.

Looking forward, several directions emerge from this work. 
Further exploration of outlier distributions and properties could deepen our understanding of their impact on model performance. 
Besides, as LFMs continue to scale, future research should focus on leveraging these insights to enhance scalability and efficiency without sacrificing performance.

We hope that this work could inspire further researches into the fundamental properties of LFMs and foster innovative approaches to address their inherent challenges, and contribute to  more robust, efficient, and generalizable large foundation models, finally paving the way for AGI.

\section*{Impact Statement}
This work offers a novel perspective on the underlying mechanisms of large foundation models (LFMs), with the potential to simplify and refine AI research. 
By investigating the intrinsic properties of LFM weights, we provide insights that could contribute to more efficient model adaptation, editing, and compression. 
Our findings propose a shift towards a more principled, physics-inspired approach, offering new avenues for understanding and improving AI models in a more systematic way.

While this research focuses primarily on LFMs, we believe the methodologies and insights presented here could have broader implications for the design and optimization of machine learning systems. 
The exploration of weight properties could lead to improvements in model robustness, interpretability, and training efficiency—key challenges in advancing AI. 
Additionally, the findings may help refine evaluation frameworks, providing a more holistic understanding of model performance beyond traditional test data.

We hope this work will contribute to a broader discourse in AI research, providing a foundation for further developments that could eventually lead to more transparent, efficient, and powerful AI systems.


\bibliography{example_paper}
\bibliographystyle{icml2025}


\newpage
\appendix
\onecolumn



\section*{Appendix}
We sincerely thank all the researchers in the community for their contributions to the advancement of the LFM field, which have deeply inspired and significantly aided the development of our work.
Due to space limitations in the main text, we will provide additional content in the appendix, which includes:
\begin{enumerate}
    \item \textbf{Exploration of Weight Distribution} (Appendix. \ref{supp weight distribution}): We provide a more in-depth exploration of weight distribution.
    \item \textbf{Open Discussion} (Appendix. \ref{sec supp open discussion}): Here, we delve deeper into the discussion of $\mathbf{W}^*$. We \textbf{strongly encourage} readers to review this section for a more profound understanding of our work.
    \item \textbf{Experimental Details} (Appendix. \ref{supp exp detail}): We include all experimental details presented in the main text.
     \item \textbf{Additional Details} (Appendix. \ref{sec supp method detail}): Other supplementary information, such as algorithm tables, will also be provided.
    \item \textbf{Related Work} (Appendix. \ref{supp related work}): We will introduce relevant literature related to our work.
    \item \textbf{Weight Distribution Plots}: We will supplement the appendix with weight distribution plots for all layers of the model.
\end{enumerate}

We hope that the researchers could have a happy journey, and we hope that after reviewing the appendix, they will gain a deeper understanding of our entire work.

\section*{Contributions}\label{sec conclusions}

The contributions of this paper are as follows:
\begin{itemize}
    \item We conducted an extensive observation and analysis of the weights in existing LFMs, uncovering distribution patterns that have never been explored in such detail before. 

    \item  We thoroughly investigate the relationship between transformation weights and Gaussian noise, drawing several significant conclusions that advance the understanding of this interaction.
    
    \item We provide an in-depth exploration of the properties of the optimal weight $\mathbf{W}^*$, leveraging these properties to explain the observed regularities in the weight distributions, and offering a more profound analysis on existing works.

    \item Building upon our findings, we propose several practical applications. Through straightforward modifications, these approaches yield exceptionally promising results.
\end{itemize}

In conclusion, we hope and believe that this work holds value and has the potential to contribute substantially to the advancement of the entire research community.

\newpage

\section{Further Exploration of Weight Distribution} \label{supp weight distribution}

\subsection{Remarks on Pre-trained Weight Distribution} \label{sec supp dis w*}

\begin{figure}[!ht]
    \centering
    \includegraphics[width=\linewidth]{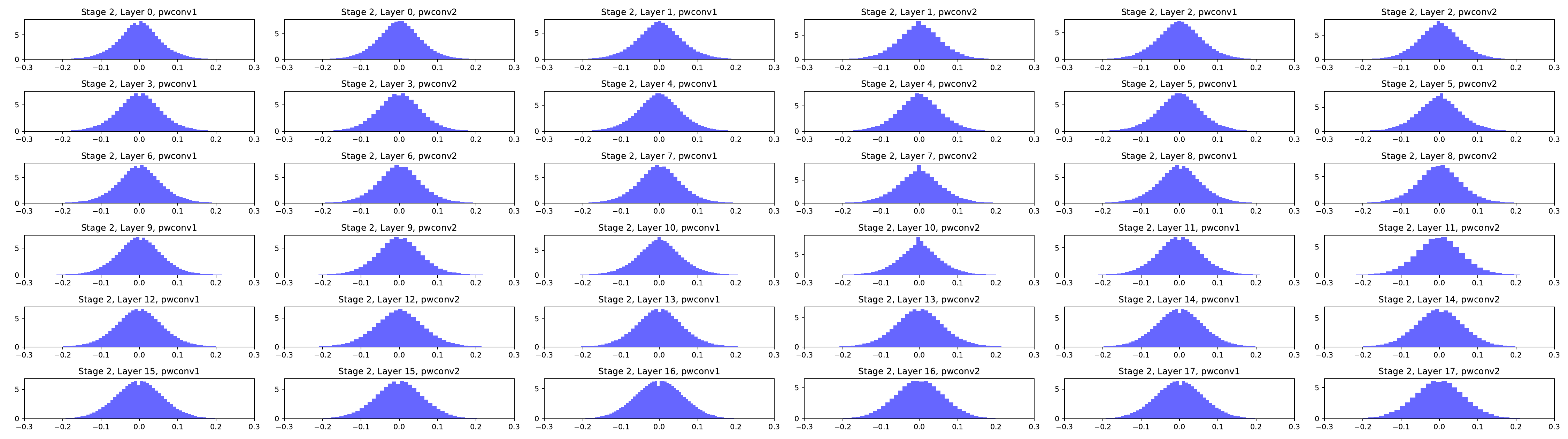}
    \caption{Weight distribution of ConvNeXt-xlarge, Stage 2, Layer 0-17, after 3$\sigma$ and extremely small value filter.}
    \label{fig:convnext filter}
\end{figure}

We here delve deeper into the distribution of the pre-trained weights.
While the majority of the weights follow a Gaussian distribution, certain weights occasionally exhibit peculiar distributions that appear unusually sharp, as shown in the first few rows of Figs. \ref{fig:qwen distribution}, \ref{fig:siglip distribution} and \ref{fig:ConvNeXt distribution}. 
These sharp distributions are primarily observed in the shallower layers of the model. 
Further analysis reveals that this sharpness arises due to a significant presence of extremely small values (magnitude smaller than 0.001) in the weight. 
We here take the weights from the 3rd-8th rows of ConvNext (stage 2, layers 0 to 17, pwconv1 and pwconv2) as an example, since they exhibit the most pronounced sharpness. 
We first validate that, disregarding the impact of very small values, the overall distribution of the weights remains Gaussian. 
Since the skewness values of these weights are still zeros, we only present the kurtosis values for these 36 layers after applying a 3$\sigma$ filter, and compare them with those obtained after further filtering out values whose magnitudes are smaller than $10^{-3}$.
The results are shown in Table. \ref{tab: kurtosis convnext}.
It can be observed that after removing the extreme small values, the kurtosis of the weights becomes closer to 3, indicating that these elements also follow a Gaussian distribution. 
We also present the distribution plots for these weights in Fig. \ref{fig:convnext filter}.

\begin{table}[!ht]
 \renewcommand\arraystretch{1}
 \setlength{\tabcolsep}{1mm}
    \centering
    \caption{Kurtosis of the weights in 36 layers in ConvNeXt-xlarge. }
    \resizebox{\textwidth}{!}{
    \begin{tabular}{ c c c c c c c c c c c c c }
    \toprule
    \multirow{2}{*}{Filter} & \multicolumn{2}{c}{Layer 0} & \multicolumn{2}{c}{Layer 1} & \multicolumn{2}{c}{Layer 2} & \multicolumn{2}{c}{Layer 3} & \multicolumn{2}{c}{Layer 4} & \multicolumn{2}{c}{Layer 5} \\ 
    & pwconv1 & pwconv2 & pwconv1 & pwconv2 & pwconv1 & pwconv2 & pwconv1 & pwconv2 & pwconv1 & pwconv2 & pwconv1 & pwconv2 \\ \midrule
    
        3$\sigma$ & 3.86 & 3.72 & 3.74 & 3.75 & 3.76 & 3.77 & 3.32 & 3.26 & 3.71 & 3.68 & 3.70 & 3.73 \\
        3$\sigma$ + $10^{-3}$ & 3.20 & 3.09 & 3.07 & 3.09 & 2.99 & 3.00 & 3.20 & 3.15 & 3.10 & 3.09 & 3.07 & 3.11 \\

        \midrule
        \midrule

        \multirow{2}{*}{Filter} & \multicolumn{2}{c}{Layer 6} & \multicolumn{2}{c}{Layer 7} & \multicolumn{2}{c}{Layer 8} & \multicolumn{2}{c}{Layer 9} & \multicolumn{2}{c}{Layer 10} & \multicolumn{2}{c}{Layer 11} \\ 
        & pwconv1 & pwconv2 & pwconv1 & pwconv2 & pwconv1 & pwconv2 & pwconv1 & pwconv2 & pwconv1 & pwconv2 & pwconv1 & pwconv2 \\ \midrule
       
       3$\sigma$ & 3.46 & 3.43 & 3.74 & 3.81 & 3.56 & 3.54  & 3.36 & 3.37 & 4.34 & 4.50 & 3.25 & 3.21 \\
        3$\sigma$ + $10^{-3}$ & 3.19 &  3.16 & 3.02 & 3.10 & 3.16 & 3.15 & 3.14 & 3.14 & 2.85 & 2.97 & 3.10 & 3.06 \\
       \midrule
       \midrule

        \multirow{2}{*}{Filter} & \multicolumn{2}{c}{Layer 12} & \multicolumn{2}{c}{Layer 13} & \multicolumn{2}{c}{Layer 14} & \multicolumn{2}{c}{Layer 15} & \multicolumn{2}{c}{Layer 16} & \multicolumn{2}{c}{Layer 17} \\ 
        & pwconv1 & pwconv2 & pwconv1 & pwconv2 & pwconv1 & pwconv2 & pwconv1 & pwconv2 & pwconv1 & pwconv2 & pwconv1 & pwconv2 \\ \midrule
        
         3$\sigma$ & 3.30 & 3.33 & 3.16 & 3.18 & 3.01 & 3.04 & 2.98 & 3.02 & 2.91 & 2.98 & 2.94 & 2.99 \\
       3$\sigma$ + $10^{-3}$ & 2.96 & 3.00 & 3.03 & 3.05 & 2.95 & 2.97 & 3.00 & 3.03 & 3.05 & 2.95 & 2.97 & 2.93 \\

    \bottomrule
    \end{tabular}
    }
    \label{tab: kurtosis convnext}
\end{table}

Indeed, similar sharp weight distributions can be observed in other models, such as Qwen2.5-32B. Upon verification, these sharp distributions are primarily caused by the presence of many small weight values, which make the overall distribution appear more ``spiked''. 
The remaining distributions can be validated, as described above, to follow a Gaussian distribution.
But why does such a distribution arise? 
A more detailed discussion will be provided in Appendix \ref{sec supp open discussion}.

\subsection{Remarks on Transformation Weight Distribution}\label{sec supp weight delta remark}

\textbf{\textit{1. Transformation Weights Follow a Gaussian Distribution:}}

We have not dedicated much effort to elaborating on the fact that the transformation follows a Gaussian distribution, since we think it is natural for the following reasons. 
Through observation, we know that both the pretrained weights and the fine-tuned weights follow a Gaussian distribution. 
Naturally, since the transformation weights are the difference between these two sets of weights, they too should follow a Gaussian distribution. 
To further elaborate it, assume that a pretrained dataset consists of $n$ samples, a downstream task dataset contains $m$ samples ($m<n$)\footnote{Generally speaking, we believe the success of adaptation is due to the knowledge acquired by the pre-trained model on a large, diverse dataset \cite{kenton2019bert,radford2021learning,brown2020language}. As a result, we typically assume that the amount of data used for pre-training is much larger than that used for fine-tuning, i.e., $m < n$.}, and both the pre-trained weights $\mathbf{W}$ and the fine-tuned weights $\mathbf{W}'$ are sampled from the optimal weights. 
Based on Eq. \ref{eq w w* gaussian}, we have
\begin{equation}
\begin{aligned}
    \mathbf{W} = \mathbf{W}^* + \mathcal{N}(0, \frac{\sigma_1^2}{n}\mathbf{I}) + o(\frac{1}{\sqrt{n}}) \\
    \mathbf{W}' = \mathbf{W}^* + \mathcal{N}(0, \frac{\sigma_2^2}{m}\mathbf{I}) + o(\frac{1}{\sqrt{m}}).
\end{aligned}
\end{equation}
Since $\mathbf{W}' = \mathbf{W} + \Delta\mathbf{W}$, we have 
\begin{equation}
    \Delta\mathbf{W} = \mathcal{N}(0, (\frac{\sigma_2^2}{m} - \frac{\sigma_1^2}{n})\mathbf{I}) + o(\frac{1}{\sqrt{m}}-\frac{1}{\sqrt{n}}).
\end{equation}
This implies that $\Delta \mathbf{W}$ follows a Gaussian distribution with a mean of 0.

The average skewness and Kurtosis of the transformation weights are shown in Table \ref{tab:skewness and kurtosis transformation}.
It is evident that as the rank increases, the distribution of the transformation weights becomes increasingly similar to a Gaussian distribution. 
This can be attributed to the fact that when the rank is small, the number of trainable parameters is limited, leading to less accurate transformation weights. 
As the rank increases, the transformation weights better approximate the true transformation weights and increasingly follow a Gaussian distribution.

\begin{table}[!ht]
 \renewcommand\arraystretch{1.05}
 \setlength{\tabcolsep}{6.8mm}
    \centering
    \caption{Average skewness and Kurtosis of the transformation weights in different settings. }
    \begin{tabular}{ c c c c c c }
    \toprule
    \multirow{2}{*}{LoRA-Dash} & \multicolumn{4}{c}{Kaiming Init.} & Xavier Init. \\ 
    & $\Delta\mathbf{W}_{r=4}$ & $\Delta\mathbf{W}_{r=8}$ & $\Delta\mathbf{W}_{r=16}$ & $\Delta\mathbf{W}_{r=32}$ & $\Delta\mathbf{W}_{r=64}$ \\ \midrule
    Ratio & 98.8\% & 99.0\% & 99.3\% & 99.0\% & 99.2\%\\
    
    Skewness & 0.00$_{\pm0.00}$ & 0.00$_{\pm0.00}$ &  0.00$_{\pm0.00}$ & 0.00$_{\pm0.00}$ & 0.00$_{\pm0.00}$ \\
    
    Kurtosis & 3.60$_{\pm0.12}$ & 3.35$_{\pm0.14}$ & 3.16$_{\pm0.14}$ & 3.33$_{\pm0.17}$ & 3.16$_{\pm0.13}$ \\
    \midrule

    \multirow{2}{*}{DoRA} & \multicolumn{4}{c}{Kaiming Init.} & Xavier Init. \\ 
    & $\Delta\mathbf{W}_{r=4}$ & $\Delta\mathbf{W}_{r=8}$ & $\Delta\mathbf{W}_{r=16}$ & $\Delta\mathbf{W}_{r=32}$ & $\Delta\mathbf{W}_{r=64}$ \\ \midrule
    
    Ratio & 98.7\% & 98.9\% & 99.1\% & 99.1\% & 99.4\% \\
    
    Skewness & 0.00$_{\pm0.00}$ & 0.00$_{\pm0.00}$ &  0.00$_{\pm0.00}$ & 0.00$_{\pm0.00}$ & 0.00$_{\pm0.00}$ \\
    
    Kurtosis & 3.75$_{\pm0.22}$ & 3.48$_{\pm0.18}$ & 3.29$_{\pm0.15}$ & 3.28$_{\pm0.15}$ & 3.07$_{\pm0.12}$ \\
    
    \bottomrule
    \end{tabular}
    \label{tab:skewness and kurtosis transformation}
\end{table}

\textit{\textbf{2. Gaussian Noise in Transformation Weights Complements the Standard Deviation of That in Pre-training:}}

Additionally, in the paper, we mention that the Gaussian noise in $\Delta \mathbf{W}$ serves to complement the standard deviation of the Gaussian noise in pre-training. 
Here, we also validate $\Delta \mathbf{W}$ obtained from LoRA-Dash and DoRA. 
We first present the results of these two methods, as shown in Table. \ref{tab:results lora-dash dora}, rows 1-2.
The best performance of each method is achieved when $r = 32$; coincidentally, their results are identical. 
Therefore, we choose $\Delta \mathbf{W}_{r = 32}$ of these two methods.

\begin{table}[!ht]
 \renewcommand\arraystretch{1}
 \setlength{\tabcolsep}{4.8mm}
    \centering
    \caption{Average results of LoRA-Dash and DoRA on commonsense reasoning tasks, and we have verified that their open-sourced weights successfully replicate the outcomes reported in their studies. 
    The upper bound (i.e., fully fine-tuning) is 81.4. }
    \begin{tabular}{ c c c c c c c }
    \toprule
    \textbf{Row} & \textbf{Methods} & $\Delta\mathbf{W}_{r=4}$ & $\Delta\mathbf{W}_{r=8}$ & $\Delta\mathbf{W}_{r=16}$ & $\Delta\mathbf{W}_{r=32}$ & $\Delta\mathbf{W}_{r=64}$ \\

    \midrule
    1 & LoRA-Dash & 75.7 & 76.9 & 75.0 & 78.4 & 76.4 \\
    2 & DoRA & 61.9 & 77.9 & 77.5 & 78.4 & 76.8 \\

    \midrule
    
    3 & $|\sigma_1-\sigma_2|$ ($\times10^{-4}$) & 54 & 45 & 81 & 2 & 35 \\
    \bottomrule
    \end{tabular}
    \label{tab:results lora-dash dora}
\end{table}

We observed that the absolute difference in standard deviations of $\Delta \mathbf{W}$ obtained from LoRA-Dash and DoRA across all layers is less than 0.0005, with an average difference of 0.0002. This shows that the standard deviations across all layers are nearly identical. This further supports the idea that the Gaussian noise in $\Delta \mathbf{W}$ complements the standard deviation of the Gaussian noise in pre-training.
It is noteworthy that these are two entirely different methods, and we are unaware of their training techniques and parameters. 
However, they exhibit phenomena that are consistent with our observations in another model, which is quite intriguing.
Moreover, their weights are publicly available, allowing interested researchers to replicate the experiments.

\textit{\textbf{3. The Closer the Standard Deviations of Two $\Delta \mathbf{W}$, the More Similar the Corresponding Model Performance Becomes:}}

Furthermore, we compared the differences in the standard deviations of $\Delta \mathbf{W}$ obtained from LoRA-Dash and DoRA for the same rank. 
The results are shown in the Table. \ref{tab:results lora-dash dora}, row 3.
Interestingly, we found that, overall, the greater the performance difference between the $\Delta \mathbf{W}$ of the same rank, the larger the difference in their corresponding standard deviations.
We thus can't help but make an intriguing hypothesis: The closer the standard deviations of $\Delta \mathbf{W}$, the more similar the corresponding model performance becomes.

\begin{figure}[!ht]
    \centering
    \includegraphics[width=\linewidth]{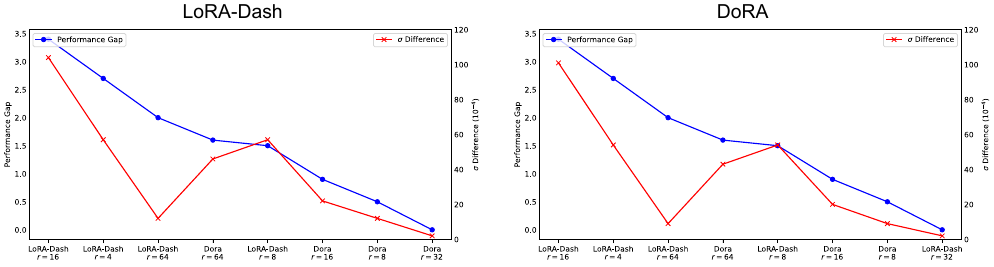}
    \caption{The relationship between $\sigma$ difference and performance gap.}
    \label{fig:sigma performance}
\end{figure}

To validate this hypothesis, we take $\Delta \mathbf{W}_{r=32}$ from both LoRA-Dash and DoRA as baselines, and compare the performance differences and standard deviation discrepancies with other settings\footnote{Given that the performance of DoRA at $r = 4$ is significantly lower, we exclude this setting from our comparison.}. 
The experimental results are presented in Fig. \ref{fig:sigma performance}, where we clearly observe that, in general, when the performance difference is smaller, the discrepancy in $\sigma$ is also smaller\footnote{We also observe that the two figures are very similar. This is because, at $r = 32$, the difference of $\sigma$ in $\Delta \mathbf{W}$ between LoRA-Dash and DoRA is already very small, resulting in negligible differences when using them as baselines.}.

Through these experiments, and considering that the role of $\Delta \mathbf{W}$ is to complement the Gaussian noise standard deviation, we believe our hypothesis is preliminarily validated. 
Furthermore, the implications of this hypothesis extend beyond its immediate context: it suggests that we can directly assess the quality of weights by comparing the standard deviation of $\Delta \mathbf{W}$. 
\textbf{This could even lead to the creation of a new evaluation system that quantifies the quality of weights without the need to evaluate performance on a test set.}
Taking PEFT as an example, we can first fully train a model to obtain an fine-tuned $\Delta \mathbf{W}$. 
Then, for any PEFT method, we can directly measure the standard deviation difference between the $\Delta \mathbf{W}$ obtained from that method and the fully fine-tuned weight, thus providing a direct quantification of the quality of the weights!

Of course, we can only say that this hypothesis is reasonable. 
However, proving it is far from possible with just our set of experiments.
We also cannot conduct all the experiments required to fully validate this hypothesis. 
Therefore, we leave this question to the researchers in the community and look forward to further studies that can verify the correctness of this conjecture in a broader context.

\textit{\textbf{4. The Standard Deviation of $\Delta \mathbf{W}$ Increases with the Layer Depth:}}

\begin{figure}[!ht]
    \centering
    \includegraphics[width=1\linewidth]{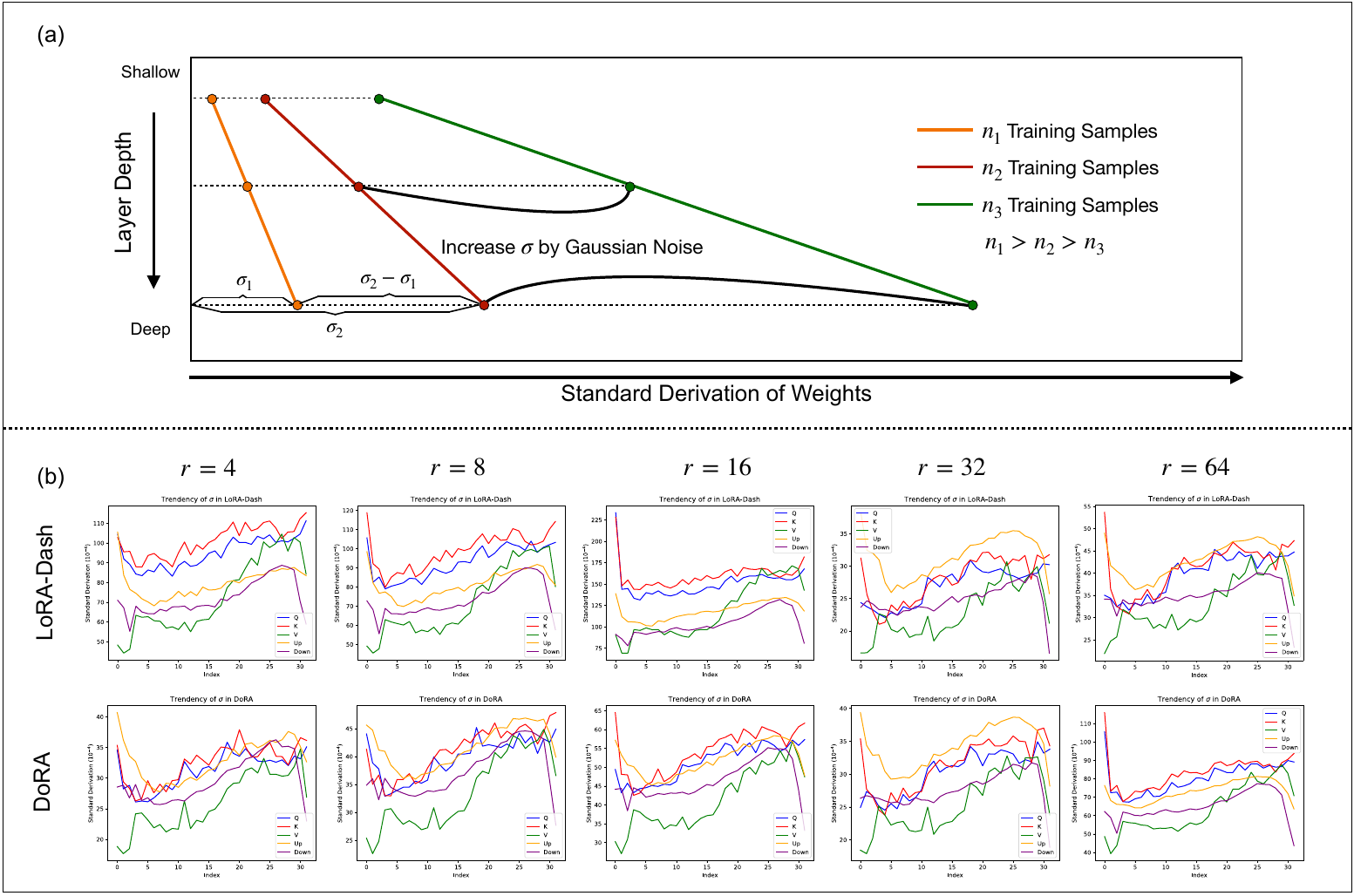}
    \caption{(a). The standard deviation of the Gaussian noise increases as the depth of the layers increases and the training data size decreases. Besides, as the training data size decreases, the increase in standard deviation becomes more pronounced in the deeper layers. (b). The variation in the standard deviation of the transformation weights trained by LoRA-Dash and DoRA.}
    \label{fig:sigma trend}
\end{figure}

As will be shown in Appendix. \ref{sec supp open discussion}, the Gaussian noise introduced during the training of each layer increases as the layer depth increases, meaning that $\sigma$ grows larger. 
Moreover, according to Eq. (\ref{eq w w* gaussian}), we know that Gaussian noise is also influenced by the size of the training data, with a larger dataset leading to a smaller $\sigma$. 
Since noise is directly related to training difficulty, we argue that when adapting a model pre-trained on a large dataset to a smaller dataset, the training difficulty across layers is not the same. 
Deep layers are inherently harder to train, and a smaller dataset only exacerbates this difficulty. 
In other words, the Gaussian noise introduced during the transfer process becomes more pronounced in deeper layers, as illustrated in Fig. \ref{fig:sigma trend}.
Therefore, the standard deviation of $\Delta \mathbf{W}$ should increase as the layer depth increases.

Therefore, we present the variation in the standard deviation of the transformation weights obtained from LoRA-Dash and DoRA across different layers in the figure.
It is clearly observed that the trend of these transformations is consistent across different layers: in most of the middle layers, the standard deviation of $\Delta \mathbf{W}$ generally increases.
We also notice that the changes in $\sigma$ in the initial and final layers are somewhat unusual, and the patterns are inconsistent across different modules. 
We speculate that this might be due to the early layers being more directly related to the data and the final layers being more task-specific.
However, the exact reason is not the focus of our current investigation.

\newpage

\section{Open Exploration on the Optimal Weights $\mathbf{W}^*$} \label{sec supp open discussion}

With a sense of both apprehension and excitement, we now begin to explore matters related to $\mathbf{W}^*$.
It is important to note that since the optimal $\mathbf{W}^*$ is unknown—and, indeed, remains unknown to the broader scientific community, obtaining and even verifying $\mathbf{W}^*$ is impossible. 
Consequently, the discussion presented here is more akin to a ``physical experiment'': based on observed phenomena, we propose plausible hypotheses and aim to validate them through a broad range of methods.

We will first propose reasonable inferences about $\mathbf{W}^*$ based on our existing observations and supports from various works in the community, and hypothesize the distribution that $\mathbf{W}^*$ should follow. 
Then, we will use the inferred distribution of $\mathbf{W}^*$ to conduct the following validations: (1) examining the causes of Gaussian and sharp distributions in the observed pretrained weights; (2) comparing with existing analyses of weight properties in the literature; and (3) evaluating the rationality of methods used in different works.

\subsection{Inferences and Conjectures Based on Existing Evidence}

First, we argue that the training process inevitably introduces zero-mean Gaussian noise. 
The larger the noise’s standard deviation, the greater its impact; conversely, a smaller standard deviation results in a lesser impact.
When the standard deviation is zero, no Gaussian noise exists. 
Many studies have indicated that, compared to deeper layers, the weights of shallow layers are more easily trained and thus more likely to approach the optimal weights \cite{lecun2002efficient, montufar2014number,choromanska2015loss}.
This suggests that the standard deviation of Gaussian noise in shallow layers is smaller than that in deeper layers, which is consistent with our observation: the standard deviation of the Gaussian distribution tends to increase across layers within the same module, as demonstrated in Fig. \ref{fig:llama distribution}.

Additionally, we find that the sharper distributions in the observed weights occur in the shallow layers—those that are well-trained and closer to the optimal weights $\mathbf{W}^*$. 
We hypothesize that when the noise's standard deviation is large, the weight distribution resembles the noise distribution, i.e., a Gaussian distribution.
Conversely, when the noise’s standard deviation is small, the corresponding Gaussian distribution becomes very ``vertical'', and the weight distribution begins to reveal more information about the distribution of $\mathbf{W}^*$.
Therefore, we will use the sharper distributions as a starting point to analyze and explore the properties of $\mathbf{W}^*$.

\textbf{First, we hypothesize that $\mathbf{W}^*$ is a sparse matrix.}
In the experiment presented in Appendix \ref{sec supp dis w*}, we observe that the sharp distribution is primarily caused by the presence of numerous small values (with magnitudes less than $10^{-3}$). These small values can account for up to 35\% of the total weights (e.g., in the ConvNext-xlarge, Stage 2, Layer 10). 
Through our review in related literature, we have noted that many approaches confirmed such small values contribute negligibly to model performance, which can safely be set to zeros \cite{yadav2024ties,deep2024della,yin2023outlier,han2015learning}. 
Based on this, we infer that these small values may stem from Gaussian noise and can be directly discarded. 
Considering that there may be many other elements that can also be set to zero, we believe that the optimal weights $\mathbf{W}^*$ should contain a significant number of zero elements, and therefore, $\mathbf{W}^*$ is a sparse matrix.
It is worth noting that the sparsity of $\mathbf{W}^*$ is consistent with findings from biological studies of brain \cite{olshausen1996emergence,barlow2001redundancy,schneidman2003synergy}.

\textbf{Second, we propose that some of the sparse values in $\mathbf{W}^*$ are outliers, i.e., elements with magnitudes significantly larger than the overall distribution.} 
This observation is both evident and direct, as we have detected outliers in all the weight distributions we have analyzed.
Moreover, many studies have emphasized the importance of outliers for model performance \cite{yin2023outlier,yadav2024ties,puccetti2022outliers}, suggesting that $\mathbf{W}^*$ indeed contains outliers.
It is also important to note that outliers could also be caused by Gaussian noise. 
In this context, we refer specifically to the outliers that exist in $\mathbf{W}^*$.

\textbf{Third, the sparse values in $\mathbf{W}^*$ follow a symmetric distribution with a zero mean.} 
We first explain why there are values other than outliers. 
We performed post-processing on the weights of LLaVA-v1.5-13B \cite{liu2024llavanext}, and when we retained only the outliers beyond 3$\sigma$ or even 2$\sigma$, the model failed to function properly. 
It produced only end-of-sequence tokens and was unable to understand context or follow instructions.
This indicates that some important values lie within the Gaussian distribution that we filtered out\footnote{Of course, a similar issue occurs if we retain only the Gaussian values within 3$\sigma$ and filter out the outliers. This suggests that $\mathbf{W}^*$ likely contains outliers.}. 
Furthermore, based on the sharp distribution, we know that part of this distribution follows a Gaussian form, which contributes to the sharpness of the distribution. Thus, when the Gaussian values are filtered out, some other values remain. Therefore, we believe that $\mathbf{W}^*$ contains values beyond just the outliers.

We then explore $\mathbf{W}^*$'s properties.
Based on our observations, we know that the mean of all pre-trained weights $\mathbf{W}$ is less than $10^{-5}$, which can be considered as zeros. 
Given the zero-mean property of Gaussian noise, and 
\begin{equation}
    \mathbb{E}(\mathbf{W}) = \mathbb{E}(\mathbf{W}^*) + \mathbb{E}(\mathcal{N}(0, \sigma^2)),
\end{equation}
we can infer that $\mathbb{E}(\mathbf{W}^*) = 0$, which means that it has a mean of zero.
Furthermore, we know that the skewness of all $\mathbf{W}$ is also zero. 
Based on the definition of skewness (details will be shown in Appendix. \ref{sec supp detail skew and kurtosis}), we have
\begin{equation}
    0 = \mathbb{E}[(\mathbf{W}-\mathbb{E}(\mathbf{W}))^3] = \mathbb{E}[(\mathbf{W}^* + \mathcal{N}(0,\sigma^2) - \mathbb{E}( \mathbf{W}^* + \mathcal{N}(0,\sigma^2) ) )^3] = \mathbb{E}[(\mathbf{W}^*-\mathbb{E}(\mathbf{W}^*))^3].
\end{equation}
Therefore, the skewness of $\mathbf{W}^*$ is also zero, indicating that it follows a symmetric distribution.

Considering the results in Appendix. \ref{sec supp dis w*}, where we found that after filtering out the extremely small values, the remaining values still nearly follow a Gaussian distribution, and taking into account the properties of $\mathbf{W}^*$, we make the following hypothesis: \textbf{the sparse values in $\mathbf{W}^*$ follow a truncated Gaussian distribution}\footnote{This is merely a hypothesis; the true distribution may differ, but this is our conjecture based on the observations.}.

Therefore, we have formulated a hypothesis about $\mathbf{W}^*$: \textit{\textbf{$\mathbf{W}^*$ is a sparse matrix, and its sparse values include a truncated Gaussian distribution as well as outliers outside this distribution. }}
Note that these properties adhere to the Occam's Razor \cite{ariew1976ockham, domingos1999role} and Newton’s principle of simplicity \cite{newton1930principia,newton1833philosophiae}.
Next, we will conduct a series of validations based on our hypothesized $\mathbf{W}^*$.

\subsection{Validation: Derivation of Weight Distribution} \label{supp derivation weight}

Based on our hypothesis about $\mathbf{W}^*$, an important question is to explain why we can observe that most weights follow a Gaussian distribution, and some follow sharp or even extreme distributions (e.g., in the Qwen2.5 model, Layer 2 MLP's Down Projection).

To investigate this phenomenon further, we first simulate $\mathbf{W}^*$.
We randomly fit a sparse matrix and assign its values a truncated Gaussian distribution. 
Specifically, we set the standard deviation of the Gaussian distribution to 0.1, and then filter out values whose absolute values are less than 0.001 or greater than 0.5, resulting in a truncated Gaussian distribution. 
Next, we simulate the training process by adding different levels of Gaussian noise to $\mathbf{W}^*$\footnote{For experimental details, please refer to Appendix. \ref{supp derivation weight}.}.

\begin{figure}[!ht]
    \centering
    \includegraphics[width=0.95\linewidth]{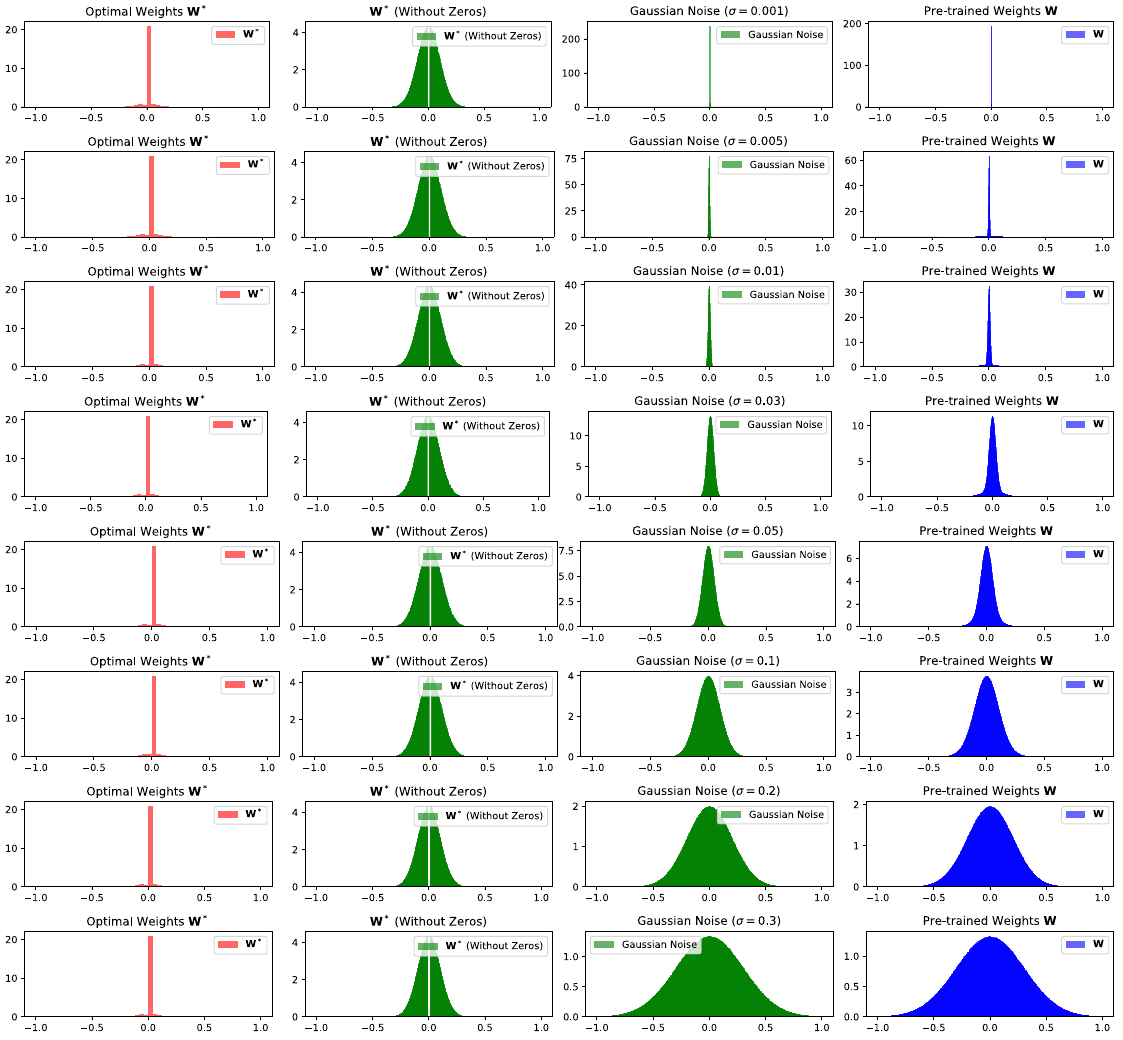}
    \caption{Derivation of Weight Distribution.}
    \label{fig:derivation weights}
\end{figure}

We present the weight distributions after adding noise at varying levels, as shown in Fig. \ref{fig:derivation weights}.
Through this figure, we gain insights into the generation of weight distributions observed in all the LFMs experiments.
We can clearly observe that
\begin{itemize}
    \item When the standard deviation of the Gaussian noise is very small (0.001, 0.005), the resulting weight distribution appears as a single line, similar to the distributions observed in Qwen2.5 Layer 1 MLP's Up Projection, Layer 2 MLP's Up Projection and Down projection etc.
    \item When the standard deviation of the Gaussian noise is relatively small (0.01), the weight distribution resembles an inverted T-shape, similar to the distributions observed in Qwen2.5 Layer 4 MLP's Up Projection, Layer 5 MLP's Up Projection, and most of the unusual distributions in ConvNext.
    \item When the standard deviation of the Gaussian noise becomes non-negligible (0.03, 0.05), the weight distribution exhibits a sharp distribution, similar to the distributions observed in Qwen2.5 Layer 2 Attn's V Projection, Layer 3 Attn's K Projection, and sharp distributions in other models.
    \item When the standard deviation of the noise is equal to or greater than the standard deviation of the optimal weights (0.1, 0.2, 0.3), the weight distribution closely resembles the Gaussian noise distribution. In this case, it aligns with the distribution observed in the vast majority of the weights.
\end{itemize}

Therefore, the Gaussian distribution observed in the vast majority of the weights is not the true distribution of the weights themselves, but rather the distribution of the noise.
Due to the large standard deviation of the noise, the true distribution of the matrix is overshadowed. 
This also explains why so many of the weights in LFM exhibit a Gaussian distribution.
When the standard deviation of the Gaussian noise is small, the true distribution of the weights begins to gradually emerge, revealing sharp distributions. 
Such distributions require a small noise standard deviation, indicating that the weights have been well-trained. 
As a result, they are usually few in number and tend to appear in the shallower layers of the model. 
Moreover, it is important to note that these sharp distributions are not the true optimal weight distributions. 
They still contain noise. 
We believe that the optimal weight distribution should resemble the one shown in the first column of Fig. \ref{fig:derivation weights}.

Therefore, based on the properties of $\mathbf{W}^*$, we have derived the weight distributions under various conditions, which correspond well to the observed cases. 
This, to some extent, validates the plausibility of our hypothesis about $\mathbf{W}^*$.

Moreover, this insight also provides an explanation for the i.i.d. properties of the weights. 
We can view the entire matrix as a mixture of Gaussian distributions, with each element being a sample from this distribution, thus exhibiting i.i.d. characteristics.

\subsection{Validation: Support from Broader Community}

We here present several works from the broader community which can be further explained by the properties of $\mathbf{W}^*$.
To begin with, we first point out several properties of $\mathbf{W}^*$: zero-mean, symmetry, and sparsity.
Next, we will explore how these properties manifest in the broader community’s methods and how they indirectly support the concept of $\mathbf{W}^*$. 
While these methods may not explicitly state that they leverage the properties of $\mathbf{W}^*$, they often use techniques that implicitly rely on these characteristics, thus providing indirect evidence for their importance.

\subsubsection{Zero-mean Property}

The property of zero mean in neural network weights, especially in deep learning models, plays a fundamental role in weight initialization, training dynamics, and generalization \cite{pascanu2013difficulty}.
It has been recognized that initializing weights with a mean of zero helps avoid symmetry-breaking issues and promotes faster convergence during training \cite{kingma2014adam}. 
Zero mean initialization is typically used in conjunction with techniques such as Gaussian distributions or uniform distributions, which ensure that the weights do not have any initial bias that could lead to skewed learning trajectories \cite{sutskever2014sequence}.

Xavier initialization \cite{glorot2010understandingtheory3} is a well-known method that sets the weights to have a zero mean and variance scaled according to the number of units in the layer, helping to maintain gradient flow and prevent vanishing or exploding gradients during training. 
This method ensures that, on average, the output of each neuron will not deviate significantly from zero, allowing the neural network to learn effectively from the start. 
Similarly, He initialization \cite{he2015delving} also assumes a zero mean for the weights, but with variance scaled according to the number of input units in the layer, which further accelerates convergence for deeper networks by maintaining appropriate gradients.

Furthermore, the assumption of zero mean in the training dynamics has been studied to understand the behavior of stochastic gradient descent (SGD). 
As pointed out by \cite{choromanska2015loss}, zero-mean weights prevent undesirable biases that could destabilize the optimization process, especially when training deep networks. 
In their work, they show that the gradient dynamics are more stable when the weights are initialized with zero mean, contributing to better optimization and generalization.

\subsubsection{Symmetry Property}

The symmetry property of neural network weights is crucial for effective learning and model expressiveness. 
Symmetric initialization prevents trivial solutions by ensuring that weights are not biased in any particular direction at the start of training. 
This is vital in deep learning, as asymmetric weight distributions can cause issues like slow convergence, poor gradient flow, and suboptimal model performance \cite{he2015delving}.
Symmetry in the weight distributions facilitates a more balanced exploration of the parameter space, allowing the model to avoid the local minima or saddle points that could otherwise hinder learning [2].

Notably, weight initialization techniques such as Xavier and He initialization are also designed to maintain symmetry, optimizing the variance of weights according to the layer's input size.
These methods are critical for deep models, where improper symmetry can lead to gradient-related problems like vanishing or exploding gradients \cite{he2016identity}.
The role of symmetry is also reflected in recent studies, which emphasize its contribution to both faster convergence and better generalization by preventing overfitting \cite{choromanska2015loss}.

\subsubsection{Sparse Property}

The sparsity property of neural network weights has gained significant attention due to its impact on both model efficiency and generalization performance. 
Sparse weight distributions are typically characterized by a large number of zero or near-zero values, which effectively reduce the model's parameter count without sacrificing performance. 
This sparsity can be viewed as a form of regularization, helping to mitigate overfitting by reducing the model's complexity \cite{han2015deep}. 
Moreover, sparse networks can lead to faster inference and reduced memory consumption, making them more suitable for deployment in resource-constrained environments \cite{wen2016learning}.

Several methods have been proposed to induce sparsity in neural networks, such as pruning, which removes unimportant weights after training, and regularization techniques like L1 regularization, which directly encourages sparsity during the optimization process \cite{tibshirani1996regression}.
The connection between sparsity and the optimal weight distribution, $\mathbf{W}^*$, is evident in many deep learning models where a large proportion of weights tend to be close to zero, aligning with the hypothesis that  $\mathbf{W}^*$ itself is sparse \cite{gale2019state}.
Furthermore, recent studies have shown that sparse models can achieve competitive performance when compared to dense models, particularly in tasks where redundancy in the model parameters is high \cite{louizos2017learning,zhu2024survey}.

\newpage

\section{Experiment Details}\label{supp exp detail}
We here report the experiment and implementation details. All the experiments were conducted on NVIDIA A100 and RTX3090 GPUs.

\subsection{Details on Large Foundation Models}\label{sec supp detail lfm}

\begin{table}[!ht]
    \centering
    \caption{Details of the LFMs.}
    \resizebox{\textwidth}{!}{
    \begin{tabular}{c c c}
    \toprule
         Model & \#Params & Project URL \\ \midrule
         LLaMA-7B \cite{touvron2023llama} & 7B & \url{https://huggingface.co/huggyllama/llama-7b} \\
         Vicuna-13B \cite{zheng2023judgingvicuna} & 13B & \url{https://huggingface.co/lmsys/vicuna-13b-v1.5} \\
        Qwen2.5-32B \cite{qwen2.5} & 32B & \url{https://huggingface.co/Qwen/Qwen2.5-32B-Instruct} \\
        SAM-h \cite{kirillov2023segment} & 641M & \url{https://github.com/facebookresearch/segment-anything} \\
        ConvNeXt-xlarge \cite{woo2023convnext} & 350M & \url{https://huggingface.co/facebook/convnext-xlarge-384-22k-1k} \\
        SigLip \cite{zhai2023sigmoid} & 428M & \url{https://huggingface.co/google/siglip-so400m-patch14-384} \\
         Idefics3-8B \cite{laurenccon2024buildingidefics3} & 8B & \url{https://huggingface.co/HuggingFaceM4/Idefics3-8B-Llama3} \\
         LLaVA-NeXT \cite{liu2024llavanext} & 8B & \url{https://huggingface.co/lmms-lab/llama3-llava-next-8b} \\
         Ovis1.6 \cite{lu2024ovis} & 3B & \url{https://huggingface.co/AIDC-AI/Ovis1.6-Llama3.2-3B} \\
         \bottomrule
    \end{tabular}
    }
    \label{tab:lfm detail}
\end{table}

We use 9 representative LFMs in our experiments. 
Specifically, 
\begin{itemize}
    \item LLaMA-7B \cite{touvron2023llama} is a large language model developed by Meta, offering a balance between model size and performance. It is designed to handle a variety of natural language processing (NLP) tasks with efficiency and scalability.
	\item Vicuna-13B \cite{zheng2023judgingvicuna} is a fine-tuned version of LLaMA-13B designed for dialogue-based tasks. It is optimized for conversational AI applications and performs well on benchmarks for open-domain question answering and dialogue generation, making it an effective model for chatbot-like applications.
	\item Qwen2.5-32B \cite{qwen2.5} is a large language model that excels in NLP tasks, such as text generation, summarization, and question answering. With 32 billion parameters, it is designed to understand and generate human-like text. 
	\item SAM-h \cite{kirillov2023segment} is a model designed for image segmentation tasks. SAM-h is a variant focused on high-resolution segmentation and the fine-grained extraction of objects from images. It is particularly useful in computer vision tasks requiring accurate and precise segmentation of complex scenes.
	\item ConvNeXt-xlarge \cite{woo2023convnext} is a CNN architecture inspired by transformers, offering state-of-the-art performance in computer vision. The xlarge version of ConvNeXt is a larger and more powerful variant, capable of handling large-scale image recognition and other visual tasks with impressive accuracy.
	\item SigLip \cite{zhai2023sigmoid} is a novel pretraining model designed for image-language tasks, which introduces a Sigmoid loss function to optimize multi-modal learning. The key innovation of SigLip lies in using Sigmoid loss, specifically for modeling the relationship between images and text, thereby enhancing joint understanding between vision and language. 
	\item Idefics3-8B \cite{laurenccon2024buildingidefics3} is a new generation of models that combines large-scale pre-training with task-specific fine-tuning for visual understanding tasks. Idefics3-8B focuses on extracting detailed visual features from images and is particularly effective in tasks such as object detection and recognition in complex visual environments.
	\item LLaVA-NeXT \cite{liu2024llavanext} is a multi-modal model that integrates language and vision for enhanced task performance. It extends the LLaMA model family by adding visual understanding capabilities, enabling it to perform cross-modal tasks like image captioning, visual question answering, and image-based reasoning.
	\item Ovis1.6 \cite{lu2024ovis} is a vision-centric model that focuses on large-scale visual data understanding. It is trained to process and understand diverse visual data types, including videos, images, and other multimedia formats. Ovis1.6 is designed for high-performance computer vision tasks such as object tracking, scene understanding, and image classification.
\end{itemize}

Strictly speaking, there are no truly ``large'' models in the CV field; we can only select the currently largest models in computer vision as a basis for exploring LFMs. 
Additionally, we chose the Vision Model of SigLiP as a foundational model for CV in our study.
The size and Project URL of each LFM are shown in Table \ref{tab:lfm detail}.

\begin{figure}[!ht]
    \centering
    \includegraphics[width=\linewidth]{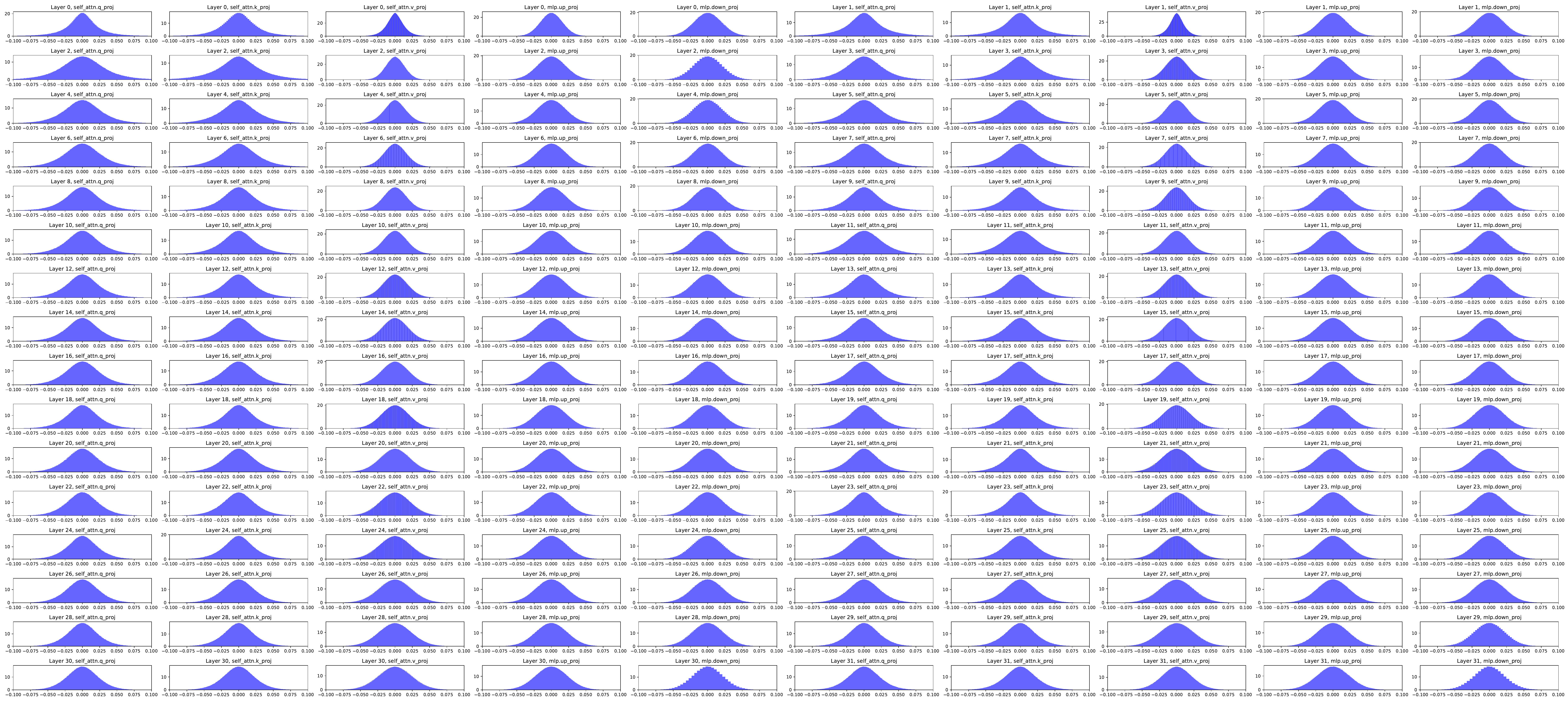}
    \caption{Weight Distribution of LLaMA-7B.}
    \label{fig:llama distribution}
\end{figure}

\begin{figure}[!ht]
    \centering
    \includegraphics[width=\linewidth]{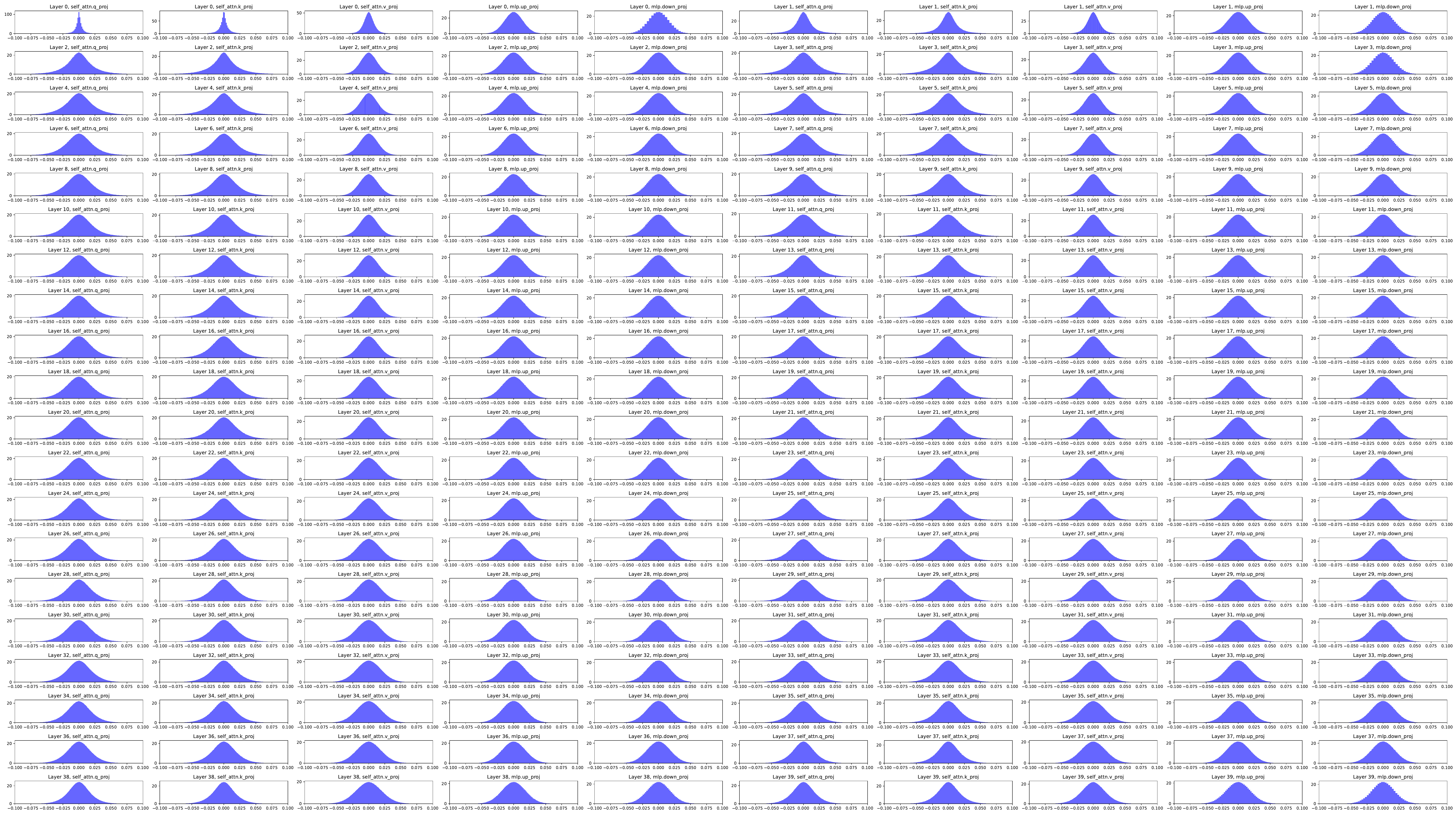}
    \caption{Weight Distribution of Vicuna-13B.}
    \label{fig:vicuna distribution}
\end{figure}

\begin{figure}[!ht]
    \centering
    \includegraphics[width=\linewidth]{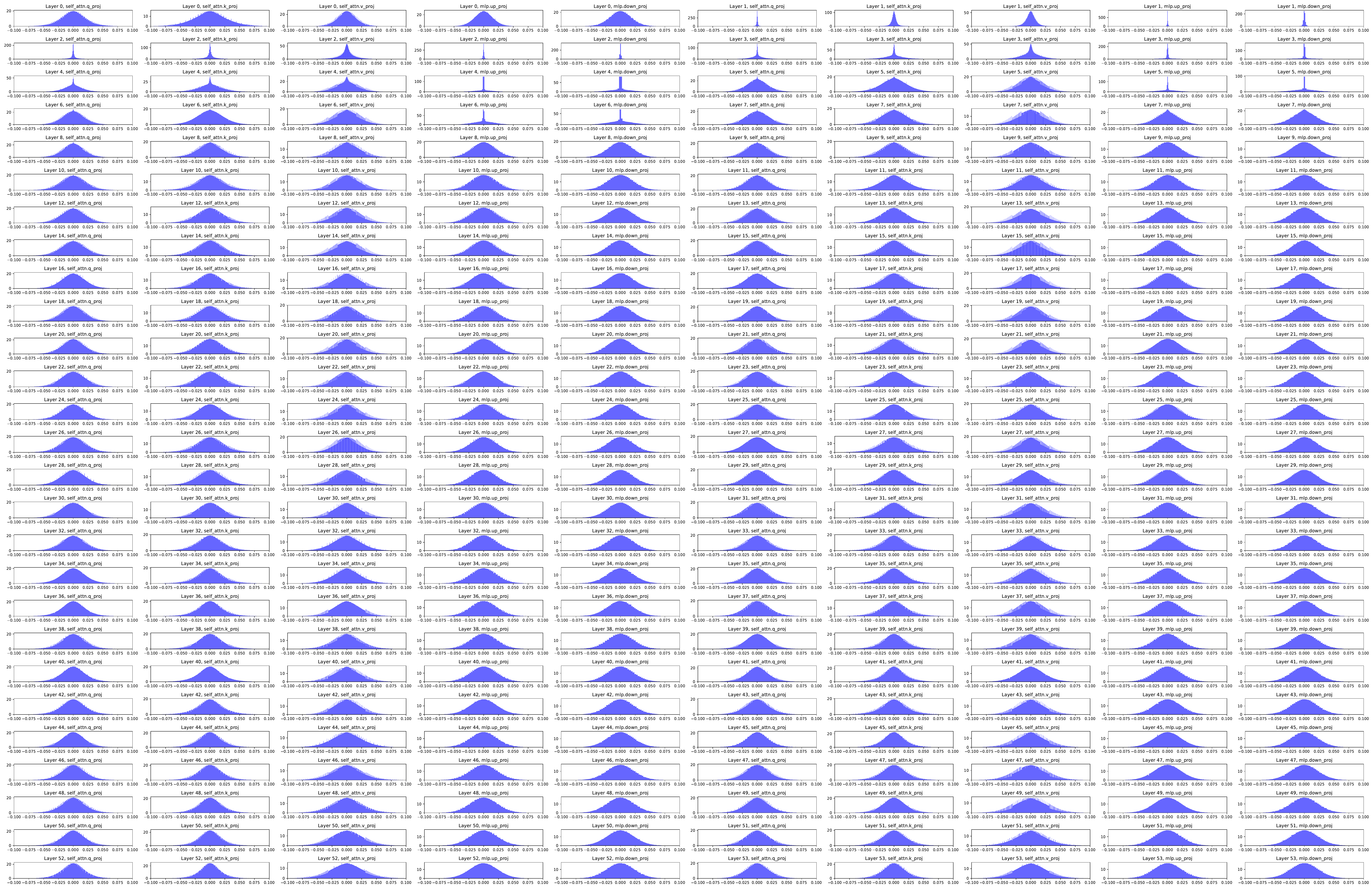}
    \caption{Weight Distribution of Qwen2.5-32B.}
    \label{fig:qwen distribution}
\end{figure}

\begin{figure}[!ht]
    \centering
    \includegraphics[width=\linewidth]{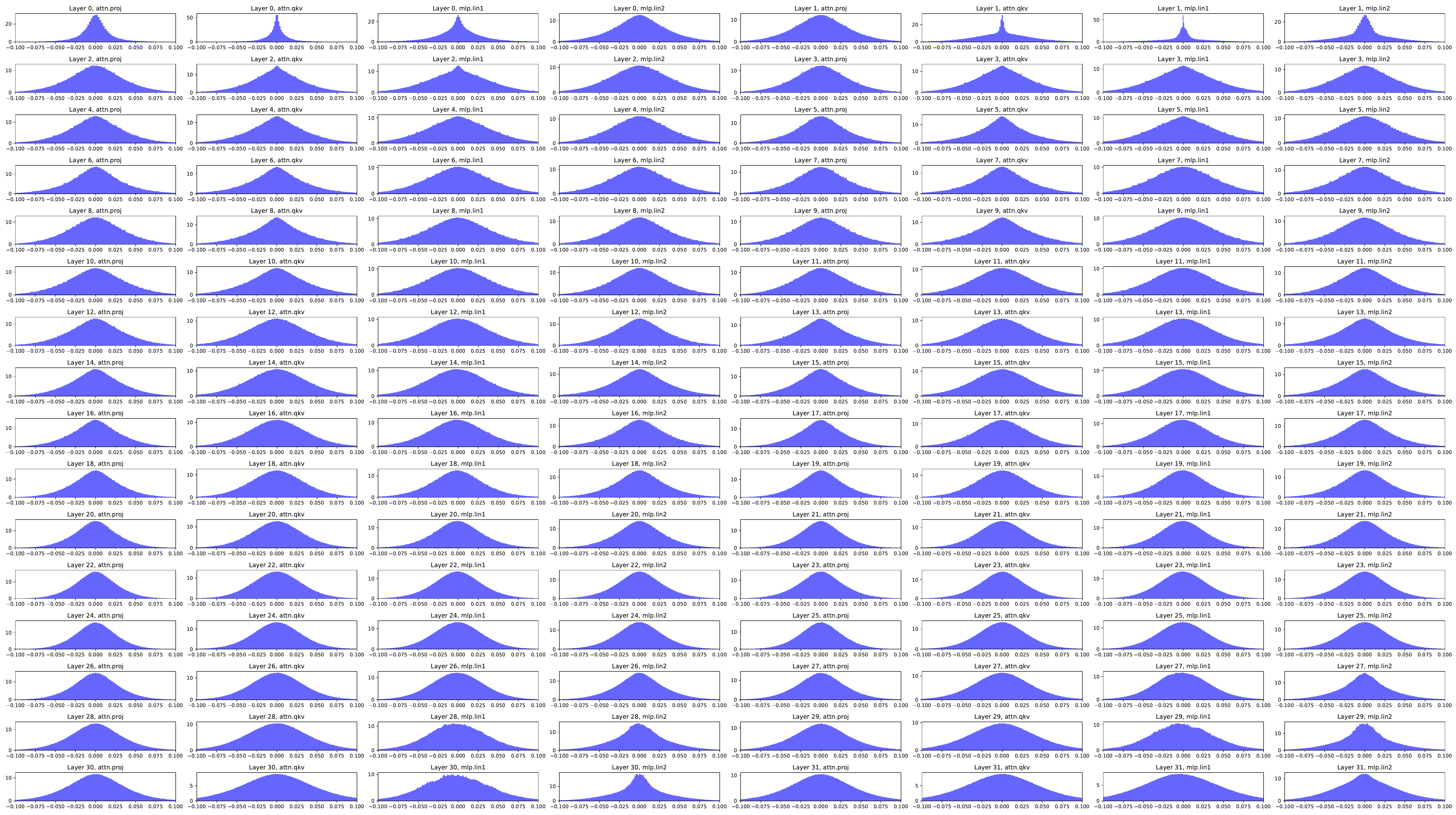}
    \caption{Weight Distribution of SAM-h.}
    \label{fig:sam distribution}
\end{figure}

\begin{figure}[!ht]
    \centering
    \includegraphics[width=\linewidth]{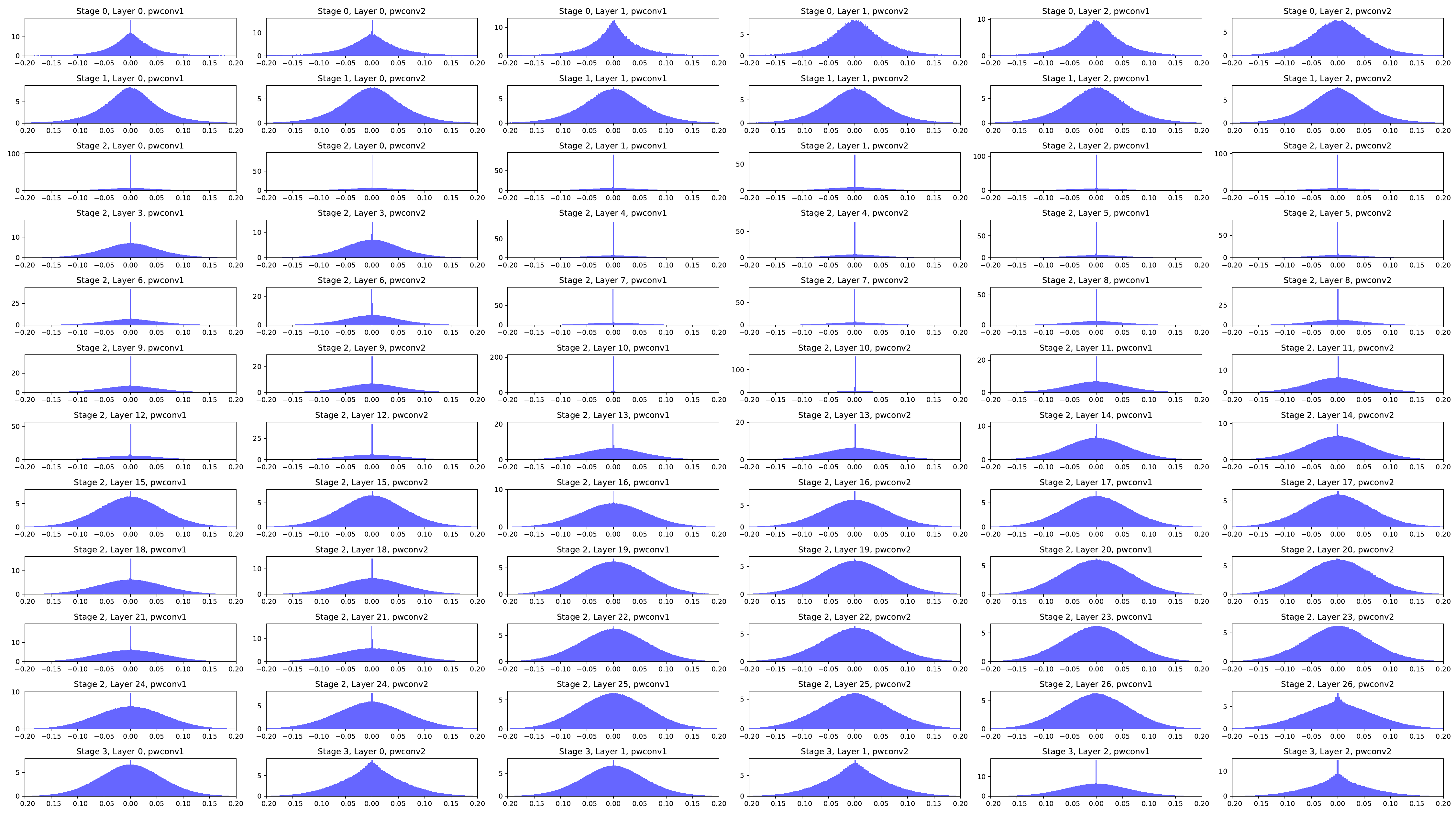}
    \caption{Weight Distribution of ConvNeXt-xlarge.}
    \label{fig:ConvNeXt distribution}
\end{figure}

\begin{figure}[!ht]
    \centering
    \includegraphics[width=\linewidth]{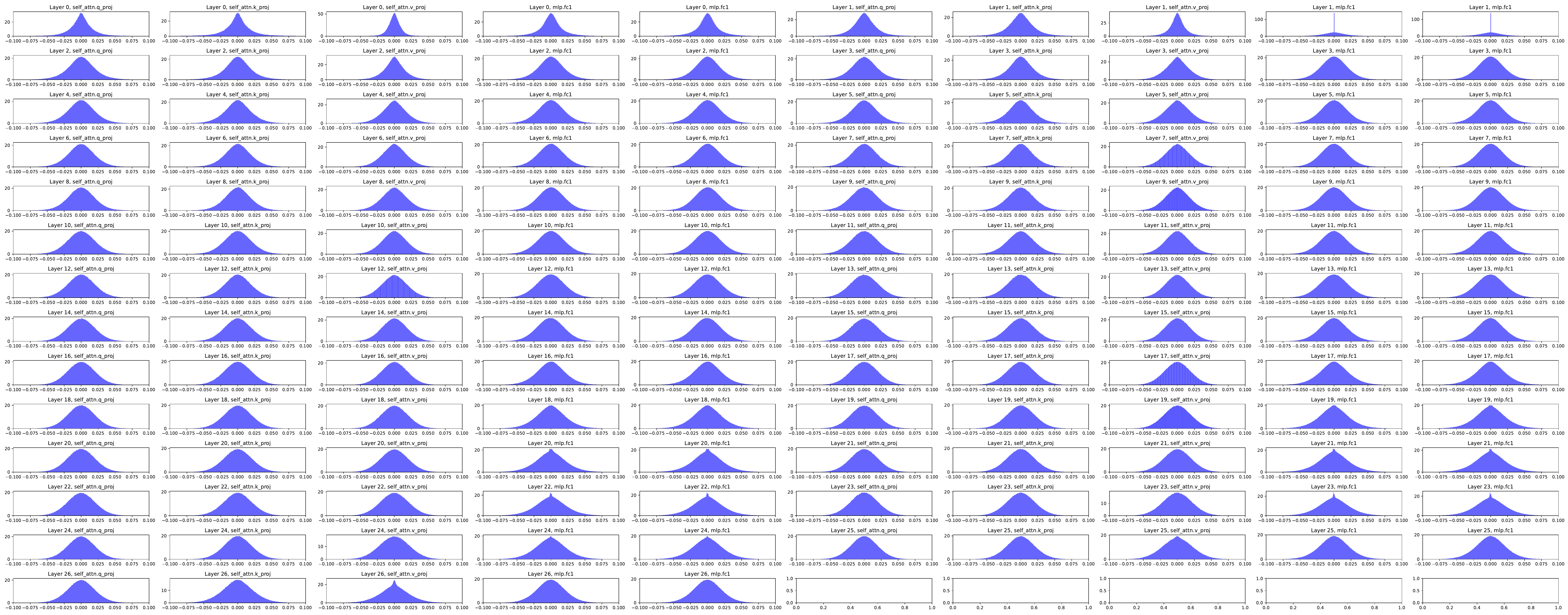}
    \caption{Weight Distribution of SigLip.}
    \label{fig:siglip distribution}
\end{figure}

\begin{figure}[!ht]
    \centering
    \includegraphics[width=\linewidth]{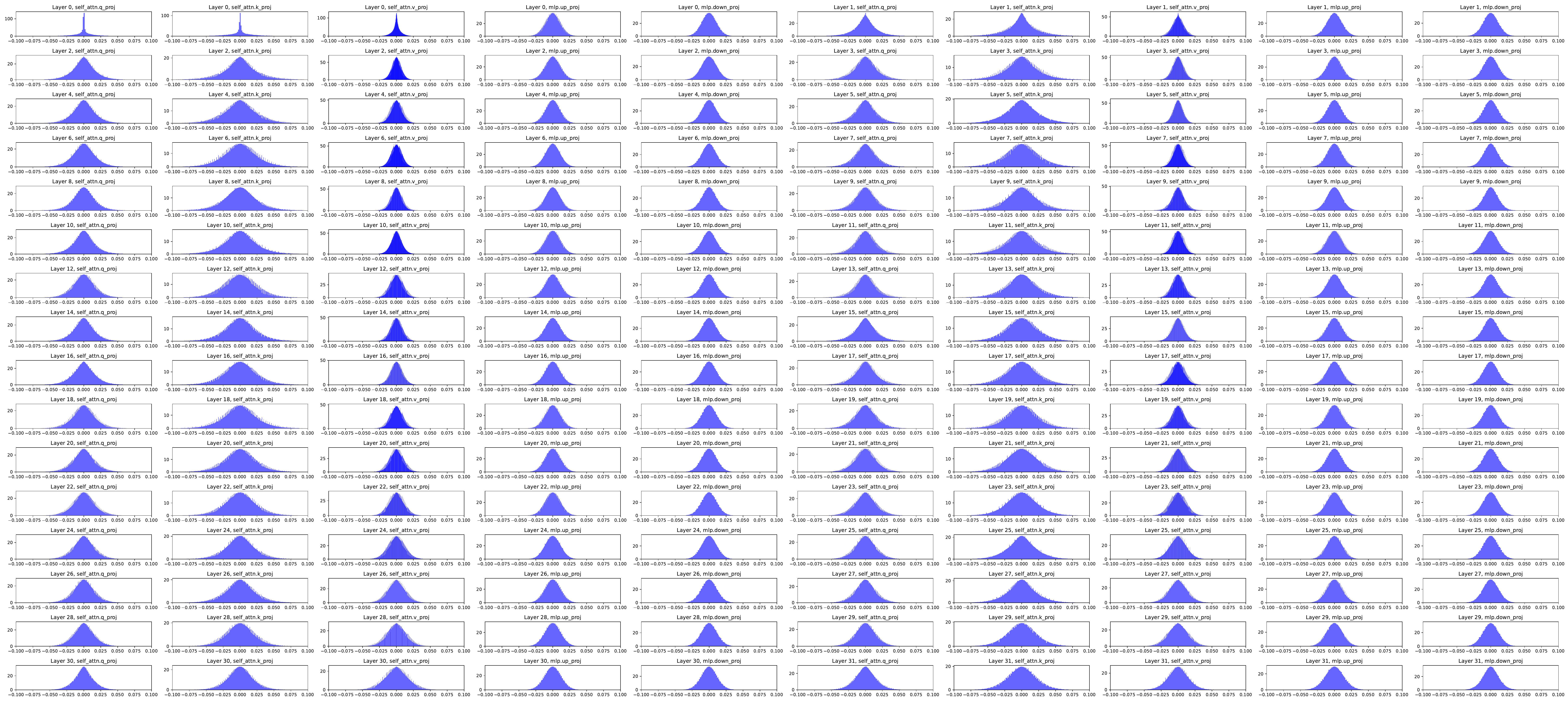}
    \caption{Weight Distribution of Idefics3-8B.}
    \label{fig:idefics3 distribution}
\end{figure}

\begin{figure}[!ht]
    \centering
    \includegraphics[width=\linewidth]{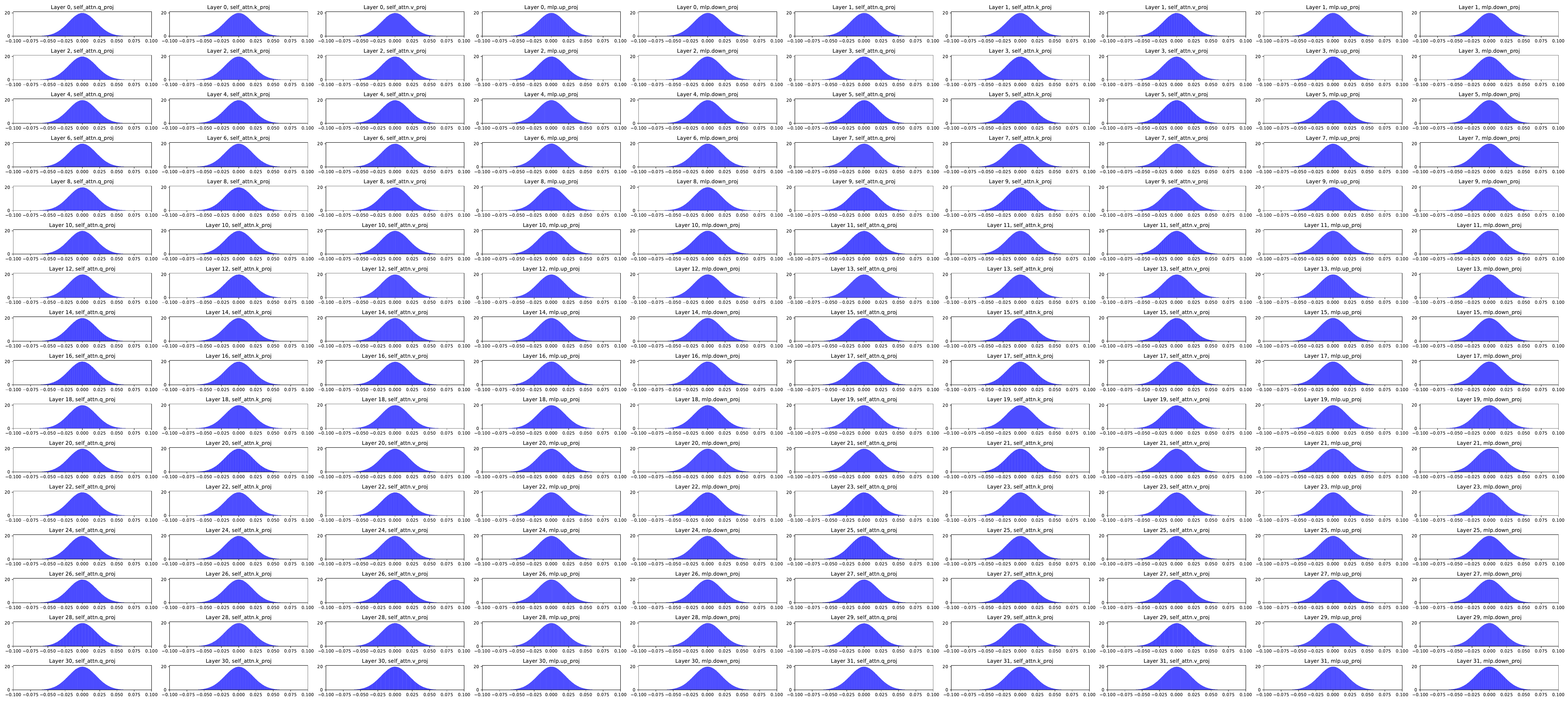}
    \caption{Weight Distribution of LLaVA-NeXT.}
    \label{fig:llava distribution}
\end{figure}

\begin{figure}[!ht]
    \centering
    \includegraphics[width=\linewidth]{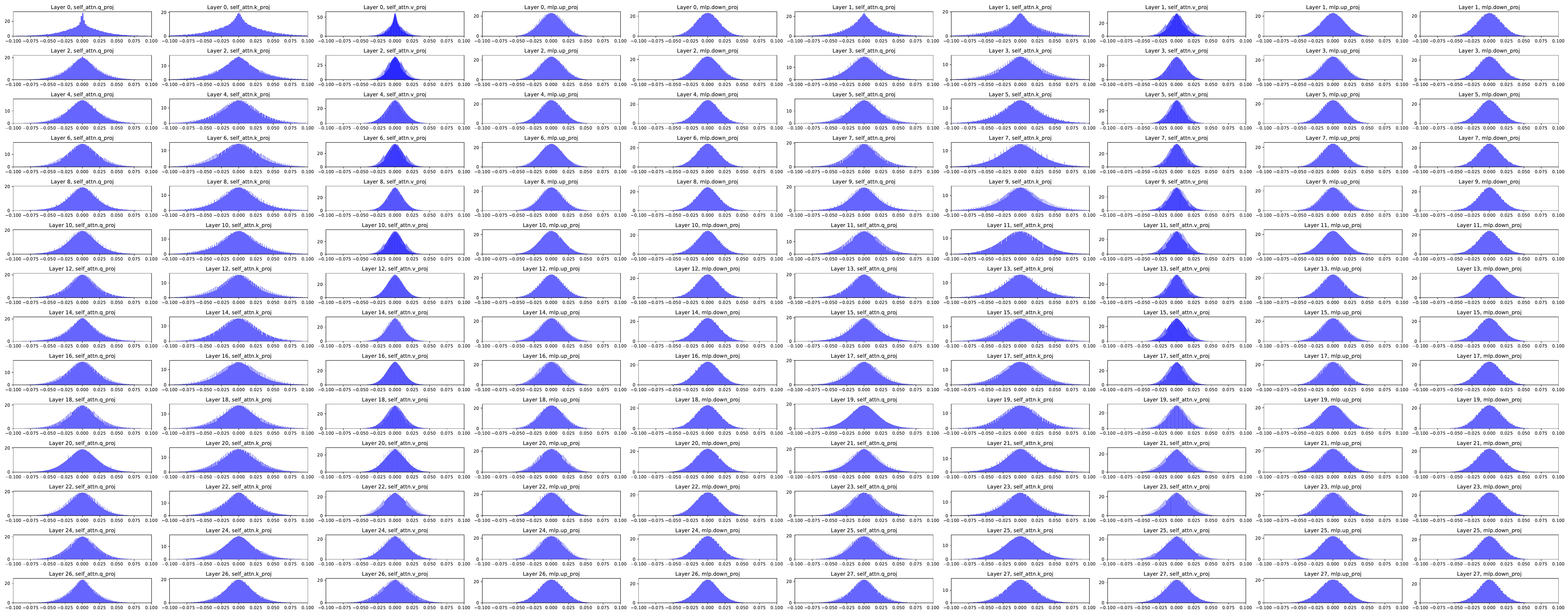}
    \caption{Weight Distribution of Ovis1.6.}
    \label{fig:ovis distribution}
\end{figure}

We also provide the weight distribution for each layer of these LFMs in Figs. \ref{fig:llama distribution}-\ref{fig:ovis distribution}.

As for the transformation weights, we use the trained $\Delta\mathbf{W}$ from LoRA-Dash\footnote{\url{https://github.com/Chongjie-Si/Subspace-Tuning}} and DoRA\footnote{\url{https://github.com/NVlabs/DoRA}}. 
Each method derives $\Delta \mathbf{W}$ through different formulations. 
In essence, they aim to learn the projection of $\Delta \mathbf{W}$ in a low-rank subspace to approximate the delta weights. The smaller the dimensions of the low-rank subspace, the smaller the trainable parameters.
In our experiments, the low-rank subspaces were set to ranks of 4, 8, 16, and 32. Furthermore, we conducted an ablation study on the initialization method with a rank of 64.
For different initialization strategies, we replace all parameters initialized with Kaiming initialization with Xavier initialization.

\begin{figure}[!ht]
    \centering
    \includegraphics[width=\linewidth]{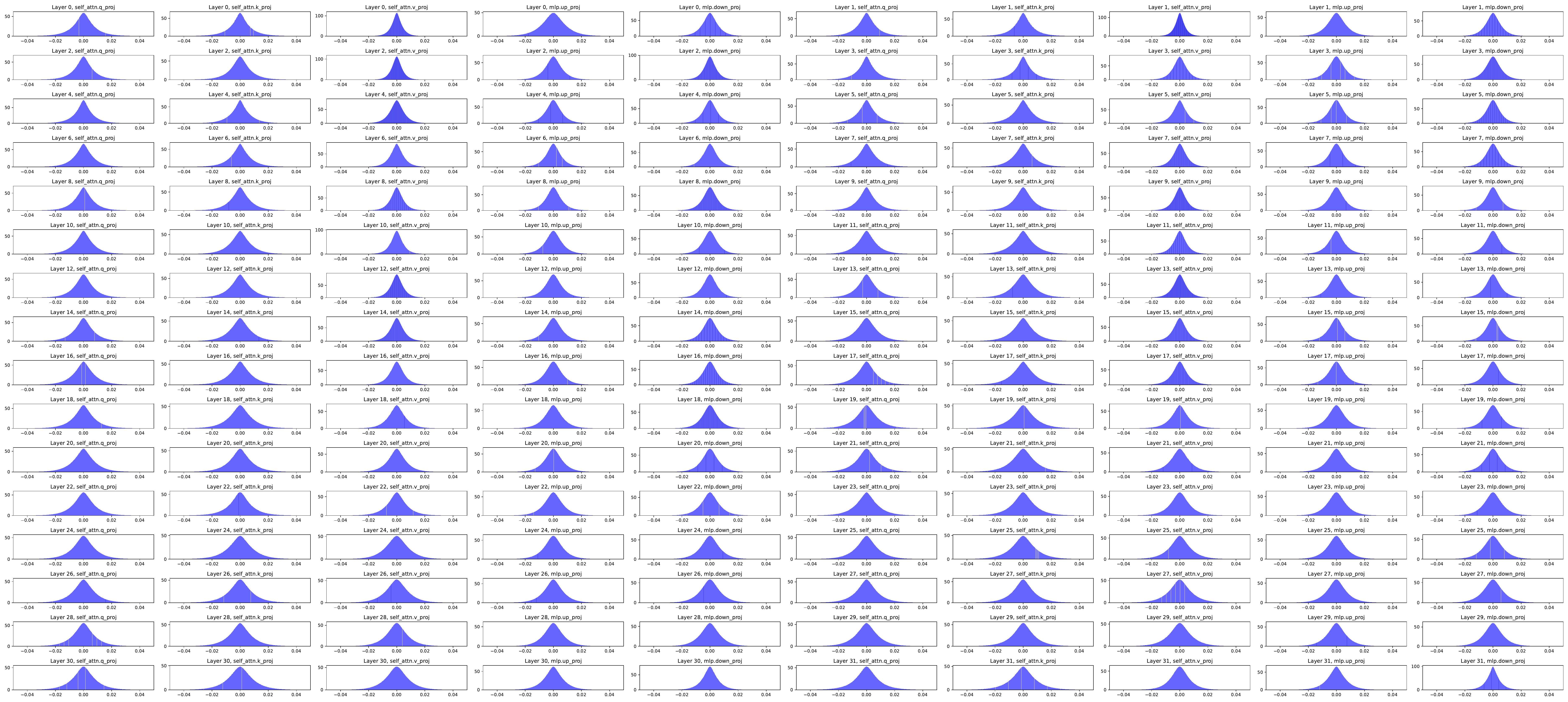}
    \caption{Weight Distribution of LoRA-Dash ($\Delta\mathbf{W}_{r=4}$) when fine-tuning LLaMA-7B.}
    \label{fig:lora-dash r=4 distribution}
\end{figure}

\begin{figure}[!ht]
    \centering
    \includegraphics[width=\linewidth]{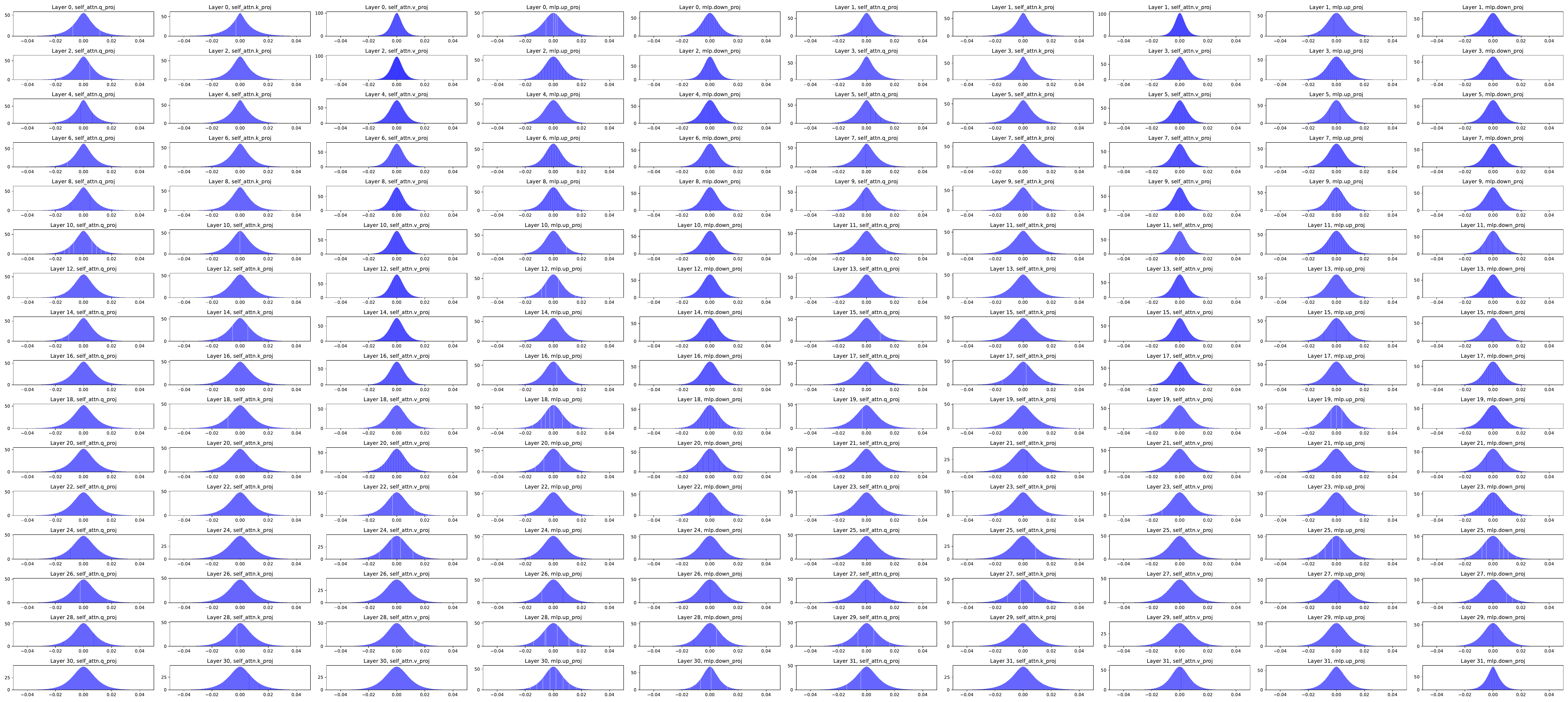}
    \caption{Weight Distribution of LoRA-Dash ($\Delta\mathbf{W}_{r=8}$) when fine-tuning LLaMA-7B.}
    \label{fig:lora-dash r=8 distribution}
\end{figure}

\begin{figure}[!ht]
    \centering
    \includegraphics[width=\linewidth]{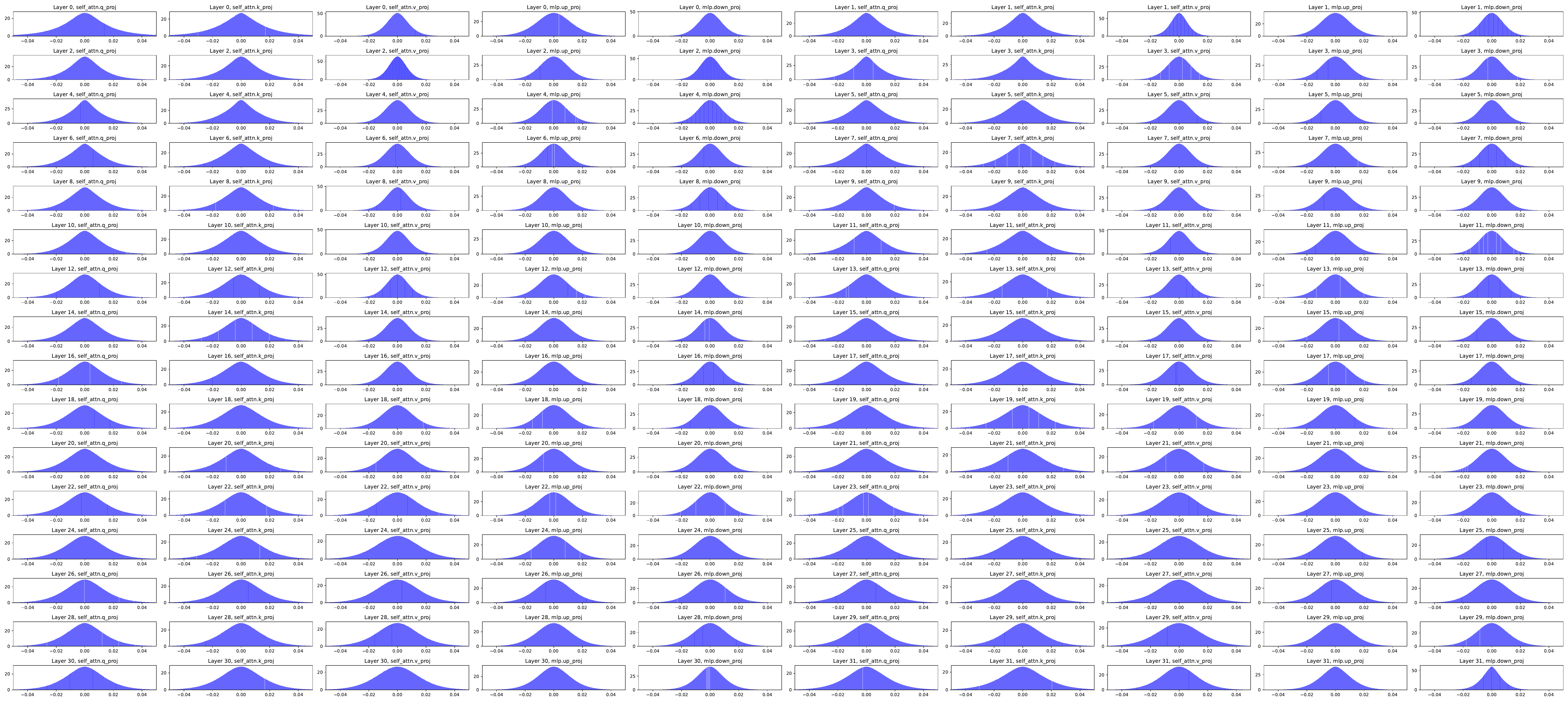}
    \caption{Weight Distribution of LoRA-Dash ($\Delta\mathbf{W}_{r=16}$) when fine-tuning LLaMA-7B.}
    \label{fig:lora-dash r=16 distribution}
\end{figure}

\begin{figure}[!ht]
    \centering
    \includegraphics[width=\linewidth]{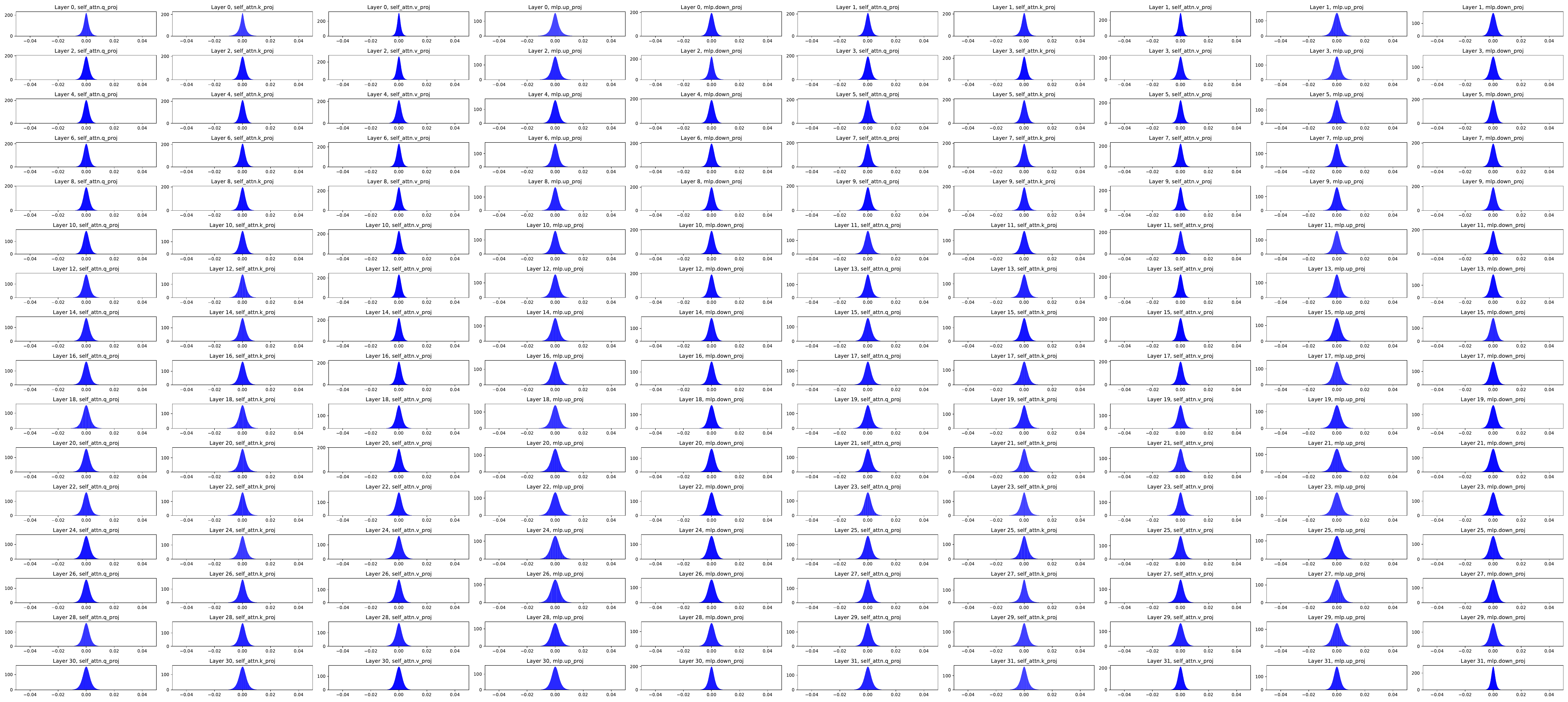}
    \caption{Weight Distribution of LoRA-Dash ($\Delta\mathbf{W}_{r=32}$) when fine-tuning LLaMA-7B.}
    \label{fig:lora-dash r=32 distribution}
\end{figure}

\begin{figure}[!ht]
    \centering
    \includegraphics[width=\linewidth]{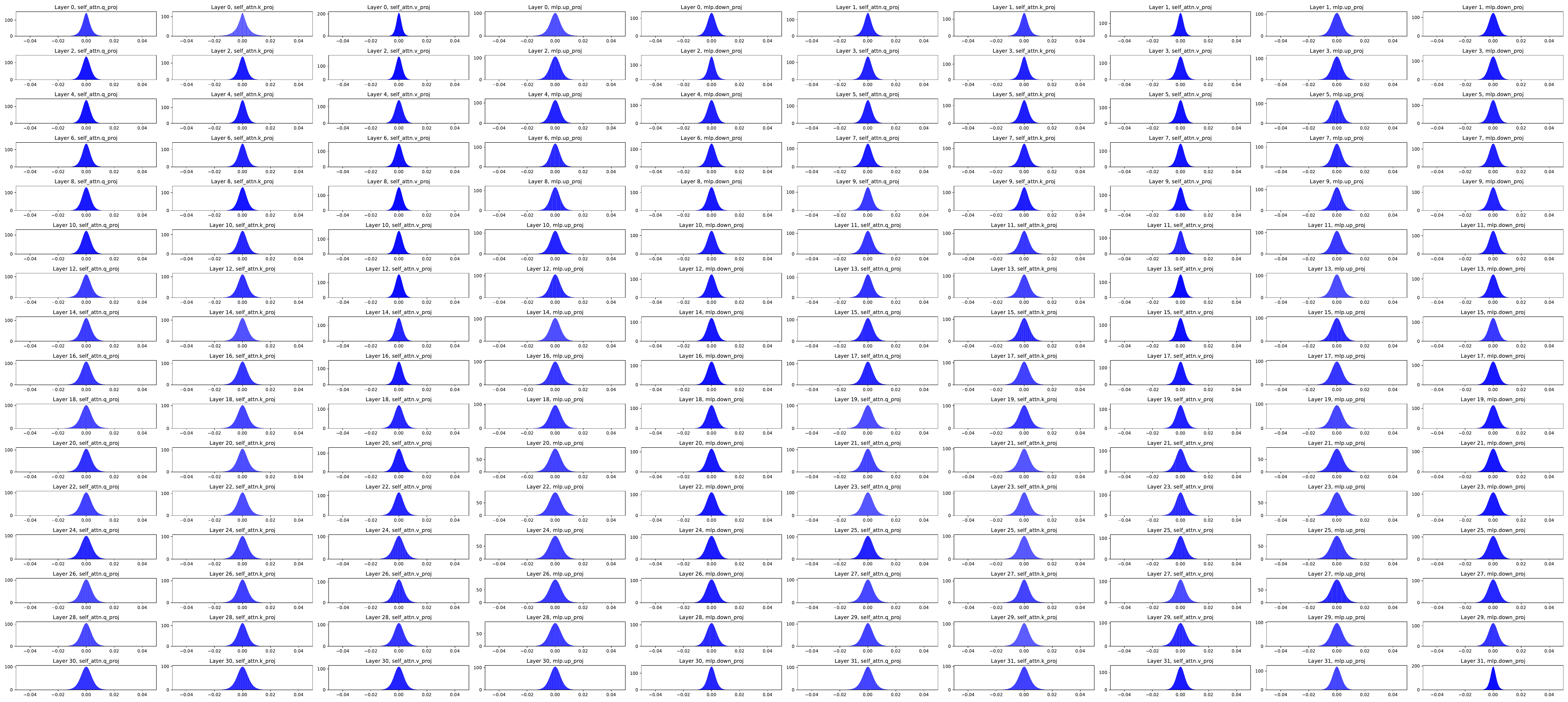}
    \caption{Weight Distribution of LoRA-Dash ($\Delta\mathbf{W}_{r=64}$) with Xavier initialization when fine-tuning LLaMA-7B.}
    \label{fig:lora-dash r=64 distribution}
\end{figure}

\begin{figure}[!ht]
    \centering
    \includegraphics[width=\linewidth]{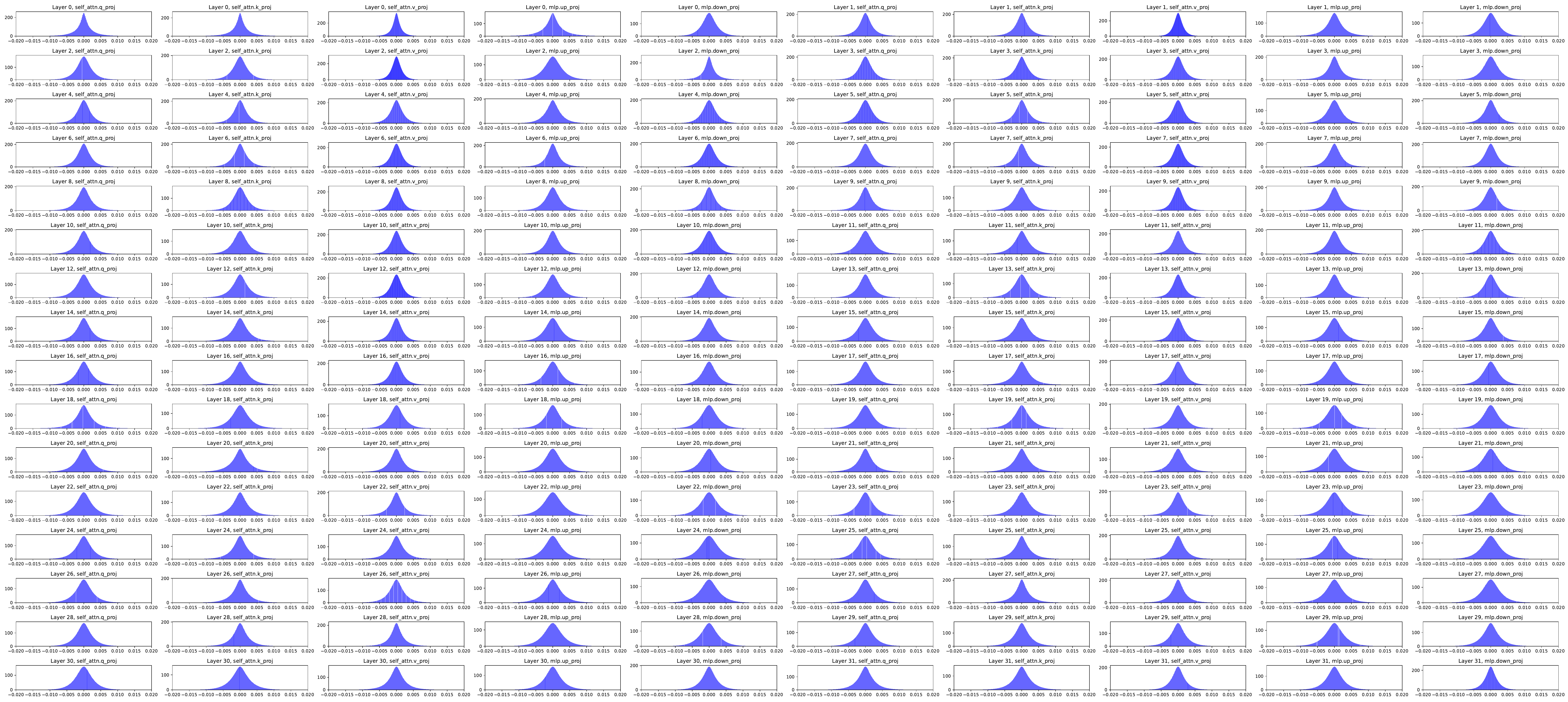}
    \caption{Weight Distribution of DoRA ($\Delta\mathbf{W}_{r=4}$) when fine-tuning LLaMA-7B.}
    \label{fig:dora r=4 distribution}
\end{figure}

\begin{figure}[!ht]
    \centering
    \includegraphics[width=\linewidth]{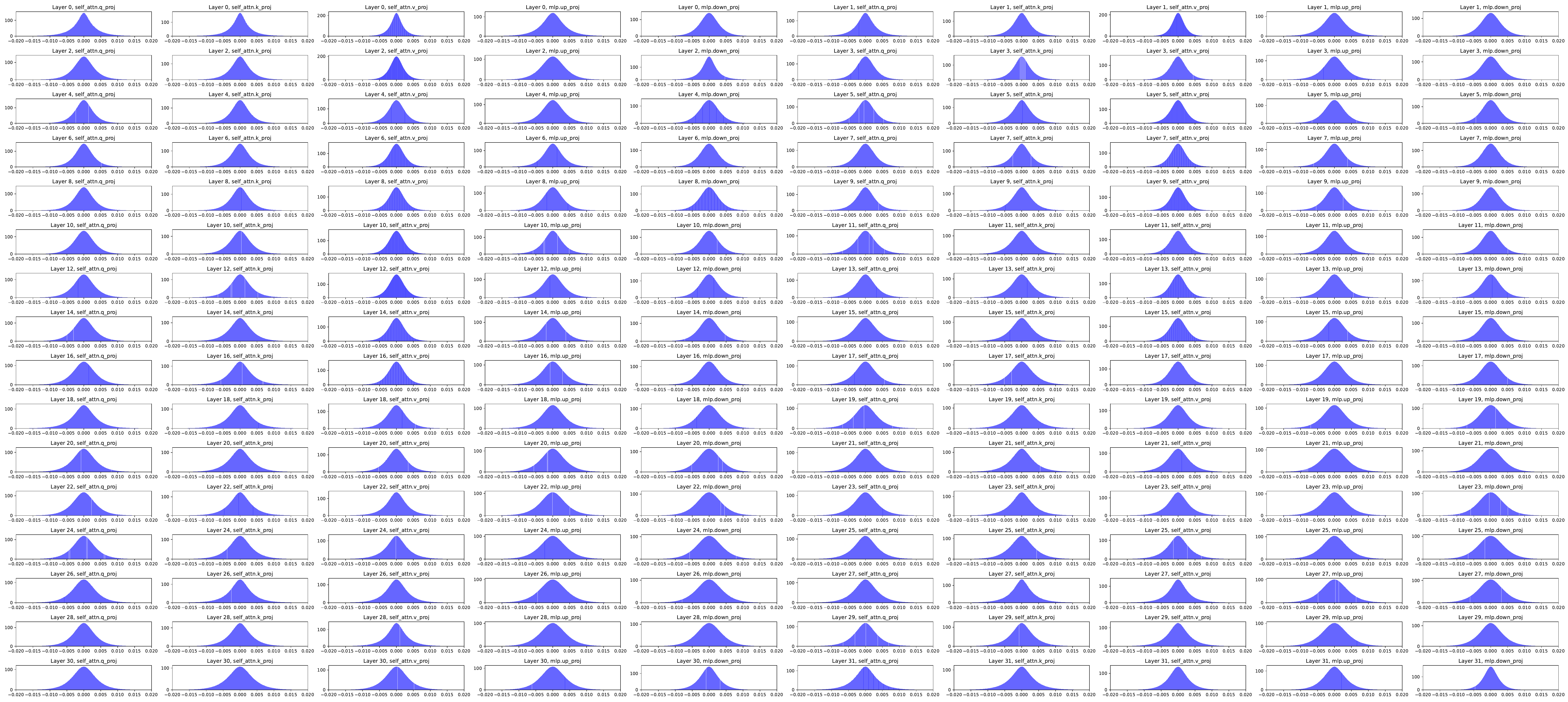}
    \caption{Weight Distribution of DoRA ($\Delta\mathbf{W}_{r=8}$) when fine-tuning LLaMA-7B.}
    \label{fig:dora r=8 distribution}
\end{figure}

\begin{figure}[!ht]
    \centering
    \includegraphics[width=\linewidth]{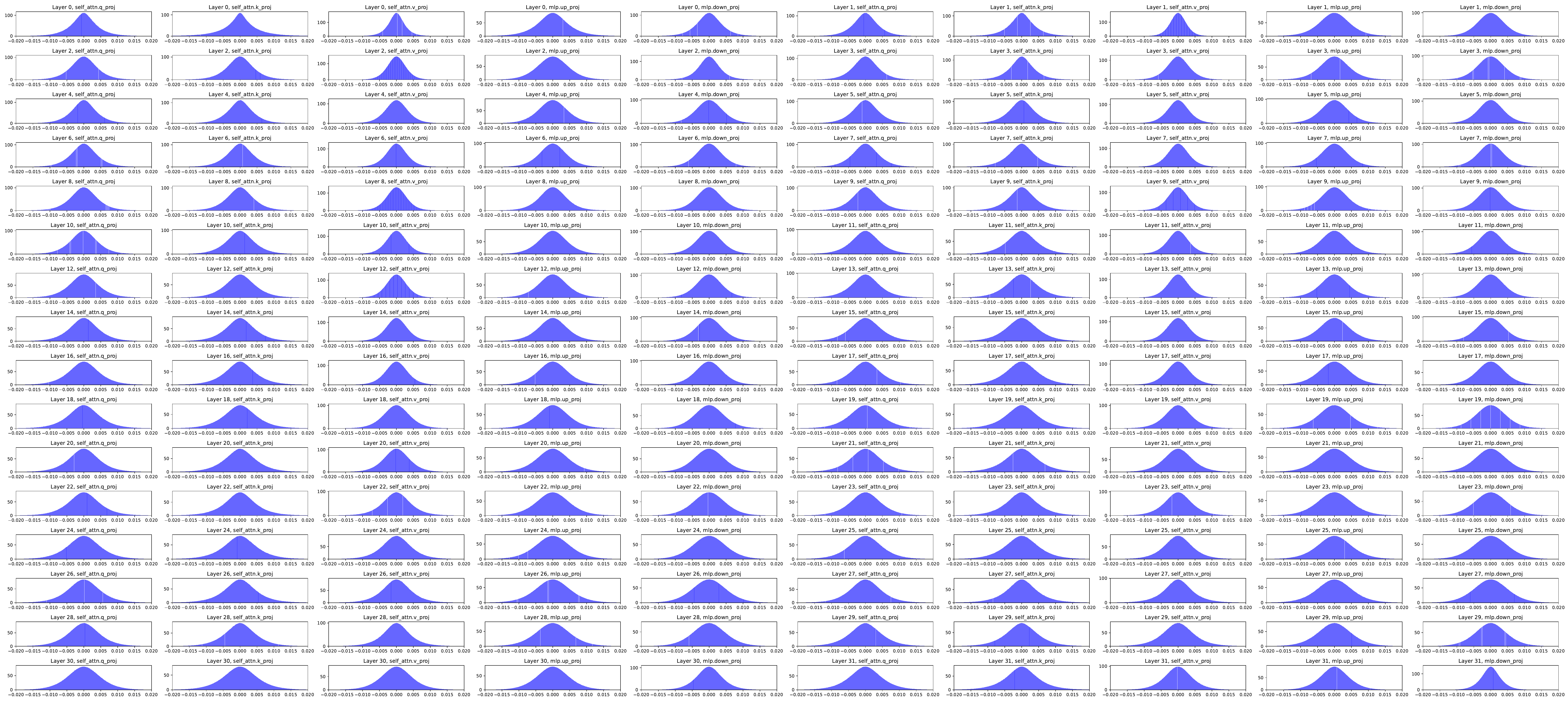}
    \caption{Weight Distribution of DoRA ($\Delta\mathbf{W}_{r=16}$) when fine-tuning LLaMA-7B.}
    \label{fig:dora r=16 distribution}
\end{figure}

\begin{figure}[!ht]
    \centering
    \includegraphics[width=\linewidth]{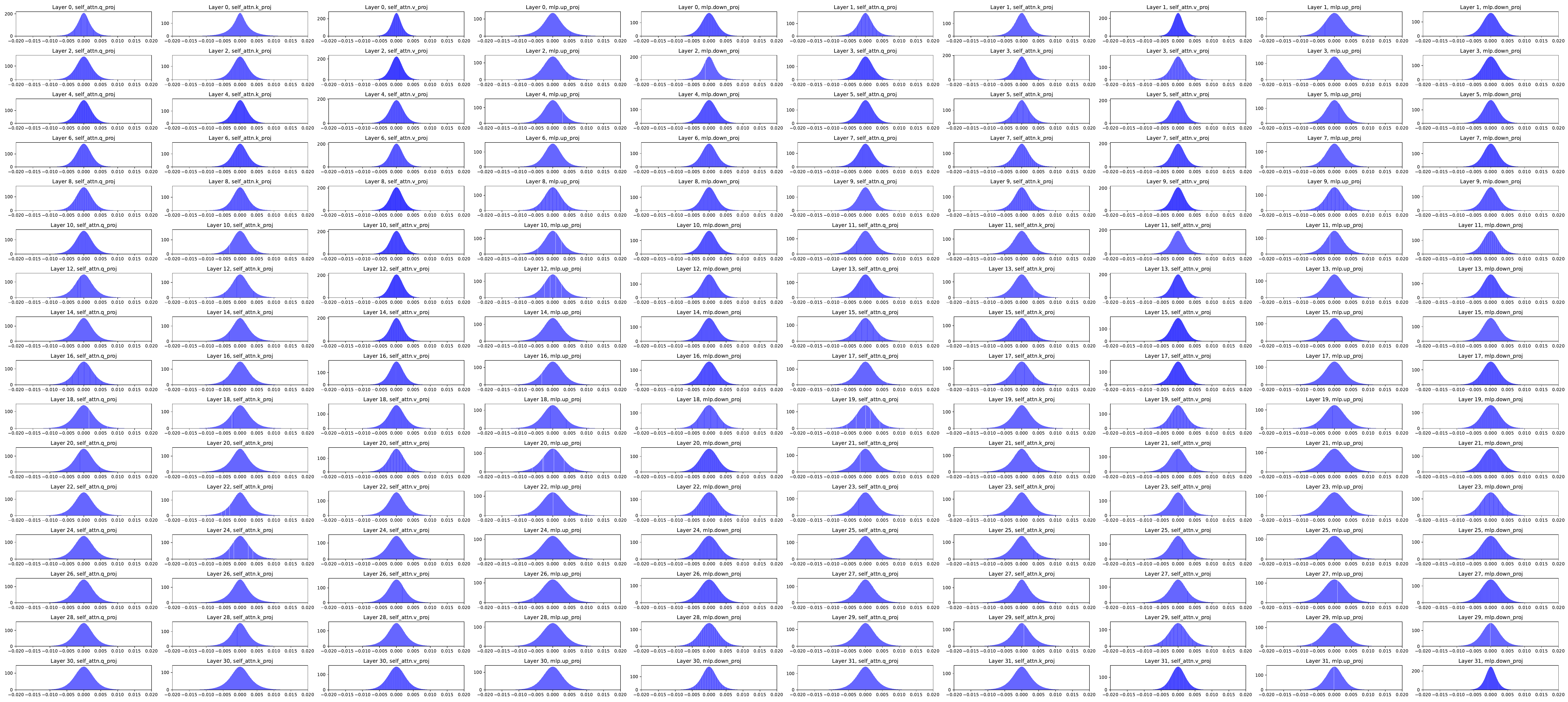}
    \caption{Weight Distribution of DoRA ($\Delta\mathbf{W}_{r=32}$) when fine-tuning LLaMA-7B.}
    \label{fig:dora r=32 distribution}
\end{figure}

\begin{figure}[!ht]
    \centering
    \includegraphics[width=\linewidth]{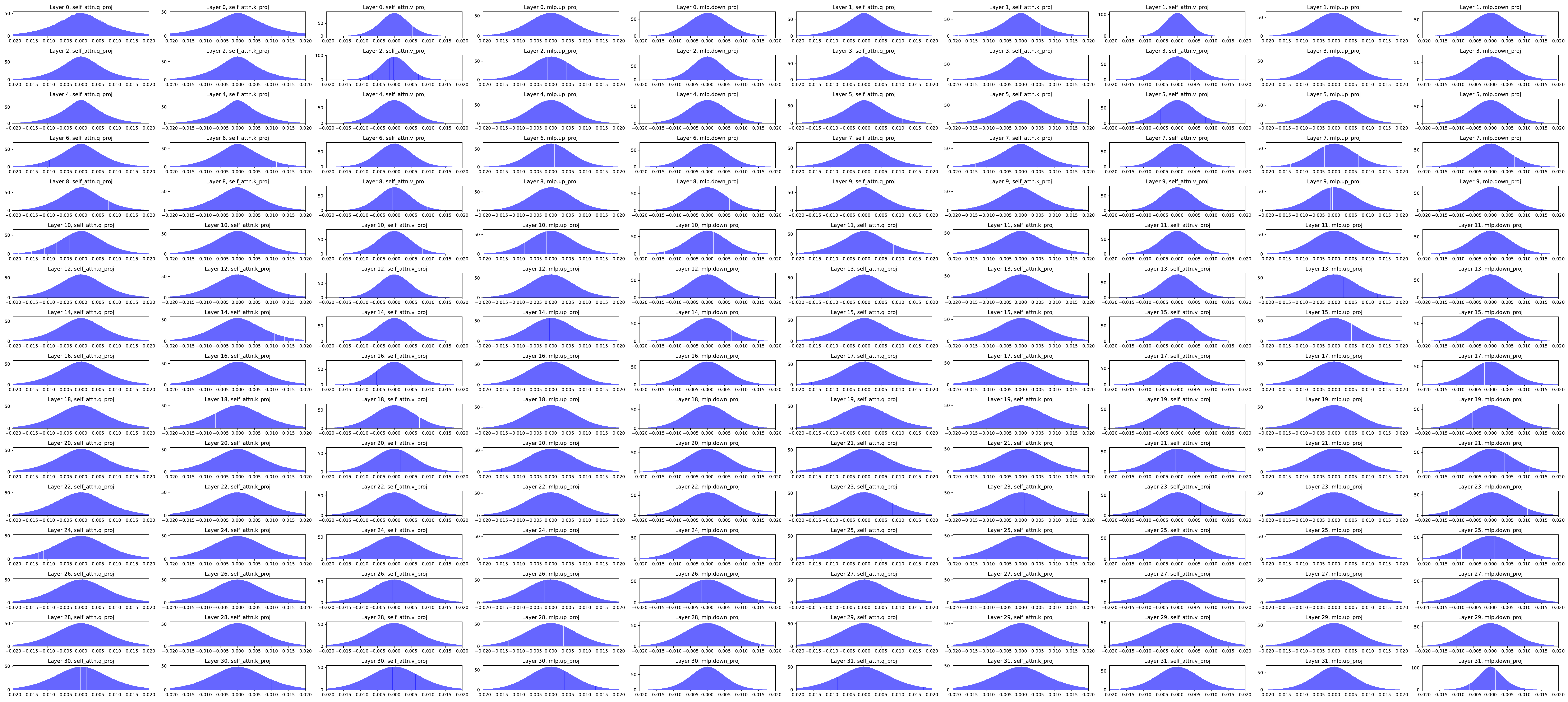}
    \caption{Weight Distribution of DoRA ($\Delta\mathbf{W}_{r=64}$) when fine-tuning LLaMA-7B.}
    \label{fig:dora r=64 distribution}
\end{figure}

We also present the distribution of transformation weights under different settings in Figs. \ref{fig:lora-dash r=4 distribution}-\ref{fig:dora r=64 distribution}.

\subsection{Details on the Skewness and Kurtosis}
\label{sec supp detail skew and kurtosis}

Skewness and Kurtosis are two metrics that can evaluate a distribution.

\textbf{Skewness} is a measure of the asymmetry of the probability distribution of a real-valued random variable. 
It indicates the degree of distortion from the normal distribution’s bell-shaped curve, with negative skewness indicating a left tail that is longer than the right tail, and positive skewness indicating a right tail that is longer than the left. 

Mathematically, skewness is defined as the third standardized moment of a distribution. The formula for skewness is:
\begin{equation}
\text{Skewness} = \frac{\mathbb{E}[(X - \mu)^3]}{(\mathbb{E}[(X - \mu)^2])^{3/2}}.
\end{equation}
where $\mu$ is the mean of the distribution and $X$ is the random variable.
For a Gaussian distribution, the skewness is obvious 0.

\textbf{Kurtosis} is a measure of the ``tailedness'' or the sharpness of the peak of the probability distribution. High kurtosis indicates a distribution with heavy tails, while low kurtosis suggests a distribution with lighter tails. 
The formula for calculating excess kurtosis is:
\begin{equation}
\text{Kurtosis} = \frac{\mathbb{E}[(X - \mu)^4]}{(\mathbb{E}[(X - \mu)^2])^2}.
\end{equation}
For a Gaussian distribution, the Kurtosis is 3.

It is important to note that other metrics, such as the Shapiro-Wilk test \cite{shapiro1965analysis} and the Kolmogorov-Smirnov test \cite{an1933sulla}, can also be used to assess whether a distribution is Gaussian. 
However, during our practical tests, we found that these metrics can sometimes provide completely contradictory conclusions for the same distribution.
For instance, distributions sampled from a standard Gaussian are often not classified as Gaussian by these tests. 
On the other hand, skewness and kurtosis have remained consistently stable in our validation process. Therefore, among these metrics, we chose skewness and kurtosis to assess Gaussianity.

\subsection{Details on the Natural Language Understanding Task in Sec. \ref{sec weight noise validation}}\label{sec: supp detail nlu task}

To evaluate performance on natural language understanding (NLU), we utilize the General Language Understanding Evaluation (GLUE) benchmark \cite{wang2018glue}, a comprehensive suite of tasks designed to assess a model’s capabilities across diverse scenarios. 
GLUE includes two single-sentence classification tasks (CoLA \cite{warstadt2019neural} and SST-2 \cite{socher2013recursive}), three similarity and paraphrase tasks (MRPC \cite{dolan2005automatically}, QQP \cite{wang2018glue}, and STS-B \cite{cer2017semeval}), and three natural language inference tasks (MNLI \cite{williams2017broad}, QNLI \cite{rajpurkar2016squad}, and RTE \cite{dagan2005pascal,bar2006second,giampiccolo2007third,bentivogli2009fifth}).
Table \ref{tab: glue dataset} provides an overview of the dataset details for each task.

\begin{table*}[!ht]
    \centering
    \caption{Details of GLUE dataset.}
    \setlength{\tabcolsep}{5mm}
    \begin{tabular}{l | l  c  c  c  c  c}
    \toprule
         Dataset & Task & \# Train & \# Dev & \# Test & \# Label & Metrics \\ \midrule
         \multicolumn{7}{c}{Single-Sentence Classification} \\ \hline
         
         CoLA & Acceptability & 8.5k & 1k & 1k & 2 & Matthews corr \\ \midrule
         
         SST-2 & Sentiment & 67k & 872 & 1.8k & 2 & Accuracy \\ \midrule
         
         \multicolumn{7}{c}{Similarity and Paraphrase} \\ \midrule

         MRPC & Paraphrase & 3.7k & 408 & 1.7k & 2 & Accuracy / F1 \\ \midrule

         QQP & Paraphrase & 364k & 40k & 391k & 2 & Accuracy / F1 \\ \midrule
         
         STS-B & Similarity & 7k & 1.5k & 1.4k & 1 & Pearson/ Spearman Corr \\  \midrule

        \multicolumn{7}{c}{Natural Language Inference} \\ \midrule
          
         MNLI & NLI & 393k & 20k & 20k & 3 & Accuracy \\ \midrule
         
         QNLI & QA/NLI & 108k & 5.7k & 5.7k & 2 & Accuracy \\ \midrule

         RTE & NLI & 2.5k & 276 & 3k & 2 & Accuracy \\
        
         \bottomrule
    \end{tabular}
    \label{tab: glue dataset}
\end{table*}

For this evaluation, we fine-tune DeBERTaV3-base model \cite{he2021debertav3} on the GLUE tasks. The specific hyper-parameter configurations used for SVDiff and Ours during fine-tuning are summarized in Table \ref{tab: nlu detail}.

\begin{table}
\centering
\caption{Hyper-parameter settings on NLU task.}
\setlength{\tabcolsep}{4mm}
\begin{tabular}{c | c c c c c c c c} 

\toprule

Dataset & MNLI & SST-2 & CoLA & QQP & QNLI & RTE & MRPC & STS-B\\ 

\midrule

Optimizer & \multicolumn{8}{c}{AdamW} \\ \midrule

Warmup Ratio & \multicolumn{8}{c}{0.1} \\ \midrule

LR schedule & \multicolumn{8}{c}{Linear} \\  \midrule

Max Seq. Len. & 256 & 128 & 64 & 320 & 512 & 320 & 320 & 128 \\ \midrule

Batch Size & 32 & 32 & 32 & 32 & 32 & 32 & 32 & 32 \\ \midrule

Learning Rate & 5e-4 & 8e-4 & 8e-4 & 1e-3 & 5e-4 & 1.2e-3 & 1e-3 & 5e-4 \\ \midrule

Epochs & 12 & 24 & 25 & 5 & 5 & 50 & 30 & 25  \\ \midrule

Module & \multicolumn{8}{c}{Q, K, V, O, Up, Down} \\

\bottomrule

\end{tabular}
\label{tab: nlu detail}
\end{table}

\subsection{Details on the Commonsense Reasoning Tasks}\label{sec supp detail cr task}

The commonsense reasoning benchmarks encompass 8 distinct sub-tasks, each associated with its respective dataset: BoolQ \cite{clark2019boolq}, PIQA \cite{bisk2020piqa}, SIQA \cite{sap2019socialiqa}, HellaSwag \cite{zellers2019hellaswag}, WinoGrande \cite{sakaguchi2021winogrande}, ARC-e/ARC-c \cite{clark2018thinkarce}, and OBQA \cite{mihaylov2018canobqa}. Following the methodology outlined in \cite{hu2023llm}, the training datasets from all sub-tasks are consolidated into a single comprehensive training dataset (Commonsense170K), and evaluations are conducted on the individual test sets of each sub-task. 

We fine-tune LLaMA-7B \cite{touvron2023llama}, LLaMA2-7B \cite{touvron2023llama2}, and LLaMA3-8B \cite{llama3modelcard} on this consolidated task. Additionally, results from ChatGPT using the gpt-3.5-turbo API are included, with particular emphasis on the zero-shot Chain of Thought approach \cite{wei2022chain}.

To ensure a fair comparison, the initial fine-tuning of models employing LoRA and LoRA+Ours is carried out using consistent LoRA configurations, with only the learning rate adjusted to achieve optimal performance. The hyper-parameter settings are presented in Table \ref{tab: cr detail}.
For our method, we initialize $s$ as zero.

\begin{table}[!ht]

\centering
\renewcommand\arraystretch{1}
\setlength{\tabcolsep}{7.5mm}
\caption{Hyper-parameter settings on commonsense reasoning task.}
\begin{tabular}{c |  c c | c c | c c} 

\toprule

\textbf{Hyper-parameters} & \multicolumn{2}{c}{LLaMA-7B} & \multicolumn{2}{c}{LLaMA2-7B} & \multicolumn{2}{c}{LLaMA3-8B}\\ 

\midrule

Rank $r$ & 16 & 32 & 16 & 32 & 16 & 32 \\ \midrule

$\alpha$ & 32 & 64 &  32 & 64 & 32 & 64 \\ \midrule

LR & 7e-5 & 3e-4 & 2e-4 & 3e-4 & 3e-4 & 3e-4 \\ \midrule

LR Scheduler & \multicolumn{6}{c}{Linear} \\ \midrule

Dropout & \multicolumn{6}{c}{0.05} \\ \midrule

Optimizer & \multicolumn{6}{c}{AdamW} \\ \midrule

Batch size & \multicolumn{6}{c}{16} \\ \midrule

Warmup Steps & \multicolumn{6}{c}{100} \\ \midrule

Epochs & \multicolumn{6}{c}{3} \\ \midrule

Where & \multicolumn{6}{c}{Q, K, V, Up, Down} \\

\bottomrule

\end{tabular}
\label{tab: cr detail}
\end{table}

\subsection{Details on the Model Merging Tasks}\label{sec supp detail mm task}

We investigate the merging of multi-modal models by evaluating the performance of different merging strategies.
Specifically, we use LLaVA-v1.5-13B \cite{liu2023llava} as the pre-trained base model.
We merge two models based on LLaVA-v1.5-13B, LLaVA-v1.6-13B \cite{liu2023llava}, a version fine-tuned on a general-purpose multi-modal dataset to improve overall comprehensive capability, and Math-LLaVA \cite{shi2024math}, a task-specific version fine-tuned for mathematical reasoning.
We test the performance on seven datasets: MathVista \cite{lu2023mathvista}, MMStar \cite{chen2024we}, MMMU \cite{yue2024mmmu}, WeMath \cite{qiao2024we}, AI2D \cite{kembhavi2016diagram}, 
DynaMath \cite{zou2024dynamath}, and GeoQA \cite{chen2021geoqa}.
We evaluate model performance using task-specific metrics, with higher scores indicating better performance. 
The average score across the three datasets is also reported as a measure of the model’s generalization ability.

\subsection{Details on the Derivation of Weight Distribution Experiment in Appendix. \ref{supp derivation weight}}

In this experiment, we have a total of 10 million points, of which only 2 million points are non-zero, satisfying the sparsity condition. 
Additionally, 0.5\% of the 2 million non-zero points are designated as outliers and assigned values in the range of [0.6, 1]. 
The remaining 99.5\% of the points are sampled from a truncated Gaussian distribution with a standard deviation of 0.1. 
We first set values with absolute magnitudes smaller than 0.001 or greater than 0.5 to zero, which results in a truncated Gaussian distribution. 
It is important to note that due to zeroing out some values, the final number of non-zero points is less than 2 million. 
Next, we generate a series of zero-mean Gaussian noise with different standard deviations, sampling a total of 10 million points. 
The values of these 10 million points are then added to the previously sparse 10 million points, producing the final observed results. 
The random seed used for the experiment is fixed at 42 to allow for reproducibility by other researchers.

\newpage

\section{Method Details}\label{sec supp method detail}

\subsection{Parameter Efficient Fine-tuning}
We provide the algorithm of our method in Alg. \ref{alg:peft}.

\begin{algorithm}[!ht]
   \caption{The pseudo code of our method in parameter efficient fine-tuning.}
   \label{alg:peft}
\begin{algorithmic}[1]
   \STATE {\bfseries Input:} Pre-trained model $\{\mathbf{W}^k|_{k=1}^p\}$, a downstream dataset $\mathcal{D}$.
   \STATE {\bfseries Output:} Trained scalar $s$ and matrices $\mathbf{A}$ and $\mathbf{B}$ for each layer.
    \STATE Initialize $s$ and $\mathbf{B}$ with zeros, $\mathbf{A}$ with Kaiming initialization.
   \FOR{epoch}
       \STATE Compute loss based on $\mathcal{D}$.
       \STATE Update $s$, $\mathbf{A}$ and $\mathbf{B}$
   \ENDFOR
   \STATE {\bfseries Return} $s$, $\mathbf{A}$ and $\mathbf{B}$.
\end{algorithmic}
\end{algorithm}

\subsection{Model Merging}
We provide the algorithm of our method in Alg. \ref{alg:model merging}.

\begin{algorithm}[!ht]
   \caption{The pseudo code of our method in model merging.}
   \label{alg:model merging}
\begin{algorithmic}[1]
   \STATE {\bfseries Input:} Pre-trained model $\{\mathbf{W}^k|_{k=1}^p\}$, $n$ fine-tuned task-specific models $\{\mathbf{W'}^k_i|_{k=1}^p\}_{i=1}^n$, hyper-parameter $t$.
   \STATE {\bfseries Output:} Merged model $\{\mathbf{W}_*^k|_{k=1}^p\}$.
   \FOR{$k=1$ to $p$}
       \STATE Compute $\{\Delta \mathbf{W}^k_i = \mathbf{W'}^k_i - \mathbf{W}^k\}_{i=1}^n$. 
       \STATE Compute $\sigma_i^k$ for each $\Delta \mathbf{W}^k_i$. 
       \STATE $\sigma^k = \min\{\sigma^k_1, \sigma^k_2, \cdots, \sigma^k_n\}$. 
    \FOR{$i=1$ to $n$}
    \STATE Update each element $\Delta w^k_{i,j}$ in $\Delta\mathbf{W}_i^k$:
    \[ \Delta w^k_{i,j} \leftarrow
    \begin{cases} 
      \frac{1}{n}\Delta w^k_{i,j}, & \text{if } \Delta w^k_{i,j} \in [ - t\sigma^k, t\sigma^k], \\
       \Delta w^k_{i,j}, & \text{if } \Delta w^k_{i,j} \notin [ - t\sigma^k, t\sigma^k].
       \end{cases}
       \]
    \ENDFOR
       \STATE  $\mathbf{W}_*^k = \mathbf{W}^k + \sum_{i=1}^n\Delta \mathbf{W}^k_i$.
   \ENDFOR
   \STATE {\bfseries Return} $\{\mathbf{W}_*^k|_{k=1}^p\}$.
\end{algorithmic}
\end{algorithm}

\newpage

\section{Related Work}\label{supp related work}

\subsection{Large Foundation Models}

Large Foundation Models (LFMs) have become a cornerstone in advancing artificial intelligence across various domains, including computer vision (CV), natural language processing (NLP), and multi-modal learning (MM). 
These models leverage extensive pre-training on massive datasets to achieve remarkable generalization capabilities, enabling them to perform effectively across a wide range of downstream tasks.

In computer vision, LFMs such as Vision Transformers (ViTs) \cite{dosovitskiy2020image} and their variants \cite{arnab2021vivit, han2022survey,liu2021swin} have significantly outperformed traditional convolutional neural networks (CNNs) \cite{he2016deep,krizhevsky2012imagenet} in tasks like object detection, semantic segmentation, and image classification. 
Moreover, multi-modal vision models, such as CLIP \cite{radford2021learning}, have bridged visual and textual domains, allowing for zero-shot and few-shot learning with minimal task-specific fine-tuning. 
Models like SAM \cite{kirillov2023segment} further demonstrate the ability of LFMs to excel in fine-grained image segmentation tasks with minimal manual intervention.

In natural language processing, models like BERT \cite{devlin2018bert}, GPT \cite{brown2020language}, LLaMA \cite{touvron2023llama}, and others \cite{team2024gemma,bai2023qwenllm,cai2024internlm2}, have set benchmarks for a variety of language understanding tasks, including question answering, sentiment analysis, and commonsense reasoning. 
These models have shown exceptional scalability and adaptability, leading to groundbreaking advancements in language generation, translation, and comprehension. 
Techniques such as self-supervised learning and attention mechanisms underpin their success, enabling the efficient encoding of complex semantic structures from massive corpora.
 
In multi-modal learning, LFMs like DALLE \cite{ramesh2021zero}, Flamingo \cite{alayrac2022flamingo}, OFA \cite{wang2022ofa}, and others \cite{liu2024llavanext,bai2023qwen,zhang2024mm1,tong2024cambrian,lu2024ovis,kondratyuk2023videopoet,zhang2023speechtokenizer,wu2023next,lu2024unified} have pushed the boundaries of cross-domain tasks by integrating visual, textual, and sometimes audio modalities. 
These models enable seamless transitions between tasks like image captioning, text-to-image generation, and multi-modal retrieval.
By leveraging shared embeddings and unified architectures, they are capable of performing complex tasks that require reasoning across multiple domains.

Despite their impressive achievements, LFMs face challenges such as computational inefficiency, memory constraints, and susceptibility to hallucination. 
Addressing these issues is crucial for their deployment across diverse real-world applications.

\subsection{Weight Distribution}

The investigation of weight distributions in neural networks has been a long-standing topic of interest, as these weights fundamentally encode the patterns learned during training. 
Early studies on weight initialization highlighted the importance of statistical properties such as Gaussian distributions for ensuring stable convergence during training \cite{he2015delving, glorot2010understandingtheory3}.
Specifically, Gaussian-distributed initializations were shown to improve optimization efficiency by maintaining balanced gradients across layers, laying the foundation for their widespread adoption.

While weight distributions in neural networks have been a topic of interest for decades, few studies have explicitly observed the Gaussian-like properties of model weights, especially in LFMs. 
Beyond initialization, \cite{anonymous2024loca} and \cite{jie2023revisiting} noted that trained weights often exhibit Gaussian-like patterns. However, existing studies have rarely validated or analyzed these observations in depth. 
Most works have either mentioned Gaussian properties in passing or treated them as incidental findings without further exploration. 
Furthermore, there has been little effort to connect these properties to the broader challenges during applications faced by LFMs, such as adaptation, editing, compression, and hallucinations.

In this context, our work fills a critical gap by systematically observing, validating, and analyzing the Gaussian nature of weights in LFMs. 
By linking this phenomenon to key challenges in LFMs, we aim to provide a unified perspective that advances the understanding of weight distributions and their role in model performance and efficiency.

\subsection{Parameter-Efficient Fine-Tuning}\label{sec related peft}
The immense complexity and computational requirements of Large Language Models (LLMs), often involving billions of parameters, pose significant challenges when adapting them to specific downstream tasks \cite{xu2023parameter,han2024parameter}. Parameter-Efficient Fine-Tuning (PEFT) has emerged as a practical solution by reducing the number of parameters and memory resources required for fine-tuning while achieving performance levels comparable to full fine-tuning \cite{si2024see,ding2023parameter}. Existing PEFT techniques can generally be categorized into three groups \cite{si2024see}: reconstruction-based methods \cite{zaken2021bitfit,fischer2024prompt}, extension-based method \cite{houlsby2019parameter,chen2022adaptformer} and combination-based method \cite{zhang2024spectral, si2024unleashing}. 
The first group enhances model performance by reconstructing the subspaces related to the singular values and vectors of the weight matrix. 
The second group focuses on introducing additional complementary subspace a new subspace and span it with the original subspace. The third group adopts both of the aforementioned mechanisms.

Research has shown that weight updates in neural networks often possess a low intrinsic rank \cite{aghajanyan2020intrinsic,li2018measuring}. 
LoRA \cite{hu2021lora}, as a predominant PEFT method, leverages this observation by parameterizing the weight updates $\Delta\mathbf{W}\in\mathbb{R}^{n\times m}$ for each layer, where the original weights are $\mathbf{W}\in\mathbb{R}^{n\times m}$, as $\Delta\mathbf{W} = \mathbf{A}\mathbf{B}$. Here, $\mathbf{A}\in\mathbb{R}^{n\times r}$ and $\mathbf{B}\in\mathbb{R}^{r\times m}$, where the rank $r$ is significantly smaller than both $n$ and $m$, enabling efficient parameter usage. For the original model output $\mathbf{h} = \mathbf{W}\mathbf{x}$, the modified forward computation becomes:
\begin{equation}
\mathbf{h} = \mathbf{W}\mathbf{x} + \Delta\mathbf{W}\mathbf{x} = (\mathbf{W} + \mathbf{A}\mathbf{B})\mathbf{x}.
\end{equation}
At the start of training, the initialization for $\mathbf{A}$ typically follows the Kaiming distribution \cite{he2015delving}, while $\mathbf{B}$ is initialized with zeros, ensuring that $\Delta\mathbf{W}$ starts at zero. During the fine-tuning process, only the low-rank matrices $\mathbf{A}$ and $\mathbf{B}$ are updated, while $\mathbf{W}$ remains fixed. At inference time, the low-rank updates are merged into $\mathbf{W}$, thereby incurring no additional computational costs. Due to its simplicity and efficiency, LoRA has gained widespread popularity.

\subsection{Model Merging}\label{sec related model merging}

Model merging techniques aim to combine multiple task-specific models into a single, unified model, preserving the strengths of each task while avoiding the need for the original training data. 
Popular methods include Average Merging \cite{wortsman2022model}, which averages model parameters, and Task Arithmetic \cite{ilharco2022editing}, which uses domain-specific offsets and scaling terms to adjust for the importance of each model. 
Fisher Merging \cite{matena2022merging} improves upon basic averaging by weighing parameters using the Fisher Information Matrix, though it comes with high computational costs. 
Other methods, such as RegMean \cite{jin2022dataless} and TIES-Merging \cite{yadav2024ties}, address parameter conflicts by optimizing regression problems and trimming low-magnitude parameters, respectively. 
Recent techniques like DARE \cite{yu2024language} reduces interference by rescaling parameters before merging. 
Model merging has various applications, including improving single-task performance, out-of-domain generalization, multitask learning, federated learning, and model compression. 

However, despite the numerous model merging methods that have been developed, a review of existing large model technical reports \cite{he2024efficient,team2024gemma,baichuan2023baichuan2} reveals that current approaches still predominantly rely on the average-based strategy. 
Based on our observations, we propose a simple yet potentially more effective strategy that outperforms the average approach.


\subsection{Model Compression}
Model compression techniques aim to reduce the size and computational requirements of machine learning models while maintaining their performance. One of the most widely used methods is pruning, which involves removing less important parameters from the model.
Techniques like weight pruning \cite{han2015learning} and neuron pruning \cite{molchanov2016pruning} identify parameters with minimal impact on the model’s output and remove them, leading to a more efficient model. 
Another common approach is quantization, which reduces the bit-width of the model’s weights. Methods such as post-training quantization \cite{jacob2018quantization} and quantization-aware training \cite{tailor2020degree} convert high-precision floating-point weights into lower-precision formats, such as int8, resulting in smaller models with faster inference times. 
Knowledge distillation \cite{gou2021knowledge} is another effective technique, where a smaller student model learns to mimic the output of a larger, pre-trained teacher model, achieving similar performance with fewer parameters.
Additionally, low-rank factorization \cite{winata2019effectiveness} approximates weight matrices as the product of two lower-rank matrices, reducing the number of parameters while preserving the model’s expressive power. 
Recently, hybrid approaches that combine pruning, quantization, and distillation have gained popularity, further improving compression efficiency without sacrificing performance. 

Upon reviewing the aforementioned methods, we identify two fundamental reasons or assumptions why a model can be compressed by these methods:
the sparsity of optimal weights, and the presence of noise in weights.
This closely aligns with our observations, and we will have a detailed discussion in Appendix. \ref{sec supp open discussion}.

\newpage

\end{document}